\documentclass[letterpaper]{article} 
\usepackage{aaai2026}  
\usepackage{times}  
\usepackage{helvet}  
\usepackage{booktabs} 
\usepackage{multirow}
\usepackage{placeins}

\usepackage{courier}  
\usepackage[hyphens]{url}  
\usepackage{graphicx} 
\usepackage{natbib}  
\usepackage{caption} 
\usepackage{tcolorbox}

\usepackage{algorithm}
\usepackage{algorithmicx}
\usepackage{algpseudocode}
\usepackage{amssymb}
\usepackage{amsmath}

\usepackage[T1]{fontenc}
\usepackage{booktabs,multirow}
\usepackage{makecell}
\usepackage{tabulary}
\usepackage{fontawesome5}

\newlength\savewidth\newcommand\shline{\noalign{\global\savewidth\arrayrulewidth
  \global\arrayrulewidth 1pt}\hline\noalign{\global\arrayrulewidth\savewidth}}

\newcommand{\tablestyle}[2]{\setlength{\tabcolsep}{#1}\renewcommand{\arraystretch}{#2}\centering\footnotesize}

\usepackage{pifont}
\usepackage{tikz}

\usepackage{algorithm}
\usepackage{bbding}
\usepackage{algpseudocode} 
\usepackage{textcomp}

\usepackage{newfloat}
\usepackage{listings}
\DeclareCaptionStyle{ruled}{labelfont=normalfont,labelsep=colon,strut=off} 
\floatstyle{ruled}
\newfloat{listing}{tb}{lst}{}
\floatname{listing}{Listing}
\title{Global Compression Commander: Plug-and-Play Inference Acceleration for High-Resolution Large Vision-Language Models}

\author{Xuyang Liu\textsuperscript{\rm 1}\thanks{This work was done during an internship at Alibaba.}, Ziming Wang\textsuperscript{\rm 2}, Junjie Chen\textsuperscript{\rm 1}, Yuhang Han\textsuperscript{\rm 3}, Yingyao Wang\textsuperscript{\rm 2}, Jiale Yuan\textsuperscript{\rm 2}, \\ 
Jun Song\textsuperscript{\rm 2$\dagger$}, Siteng Huang\textsuperscript{\rm 4}, Honggang Chen\textsuperscript{\rm 1, \rm 5}\thanks{Corresponding author.}
}
\affiliations {
{$^1$ Sichuan University}, \\
{$^2$ Taobao \& Tmall Group of Alibaba}, \\{$^3$ Westlake University},\\ 
{$^4$ Zhejiang University}, \\
{$^5$ Police Integration Computing Key Laboratory of Sichuan Province} \\
liuxuyang@stu.scu.edu.cn, honggang\_chen@scu.edu.cn}

\begin{document}

\maketitle

\begin{abstract}

Large vision-language models (LVLMs) excel at visual understanding, but face efficiency challenges due to quadratic complexity in processing long multi-modal contexts. While token compression can reduce computational costs, existing approaches are designed for single-view LVLMs and fail to consider the unique multi-view characteristics of high-resolution LVLMs with dynamic cropping. Existing methods treat all tokens uniformly, but our analysis reveals that global thumbnails can naturally guide the compression of local crops by providing holistic context for informativeness evaluation. In this paper, we first analyze dynamic cropping strategy, revealing both the complementary nature between thumbnails and crops, and the distinctive characteristics across different crops. Based on our observations, we propose ``Global Compression Commander'' (\textit{i.e.}, \textbf{GlobalCom$^2$}), a novel plug-and-play token compression framework for HR-LVLMs. GlobalCom$^2$ leverages thumbnail as the ``commander'' to guide the compression of local crops, adaptively preserving informative details while eliminating redundancy. Extensive experiments show that GlobalCom$^2$ maintains over \textbf{90\%} performance while compressing \textbf{90\%} visual tokens, reducing FLOPs and peak memory to \textbf{9.1\%} and \textbf{60\%}.


\end{abstract}

\begin{links}
    \link{Code}{https://github.com/xuyang-liu16/GlobalCom2}
    \link{Extended version}{https://arxiv.org/abs/2501.05179}
\end{links}
\section{Introduction}
\label{sec:intro}

By bridging visual encoders with large language models (LLMs)~\cite{Touvron:LLaMA,yang2024Qwen2}, large vision-language models (LVLMs)~\cite{Liu:LLaVA-1.5,Bai:Qwen-VL} have recently achieved remarkable progress. As LVLMs advance towards high-resolution image understanding (HR-LVLMs), dynamic cropping has emerged as a de facto standard, represents a high-resolution image as a global thumbnail combined with a set of local crops. While this enables models like LLaVA-NeXT~\cite{Liu:LLaVA-NeXT} and InternVL 3~\cite{zhu2025internvl3} to capture fine-grained details with vision encoding efficiency, it also introduces challenges with increased visual tokens and hierarchical visual context.

\begin{figure}[!t]
  \centering
  \includegraphics[width=\linewidth]{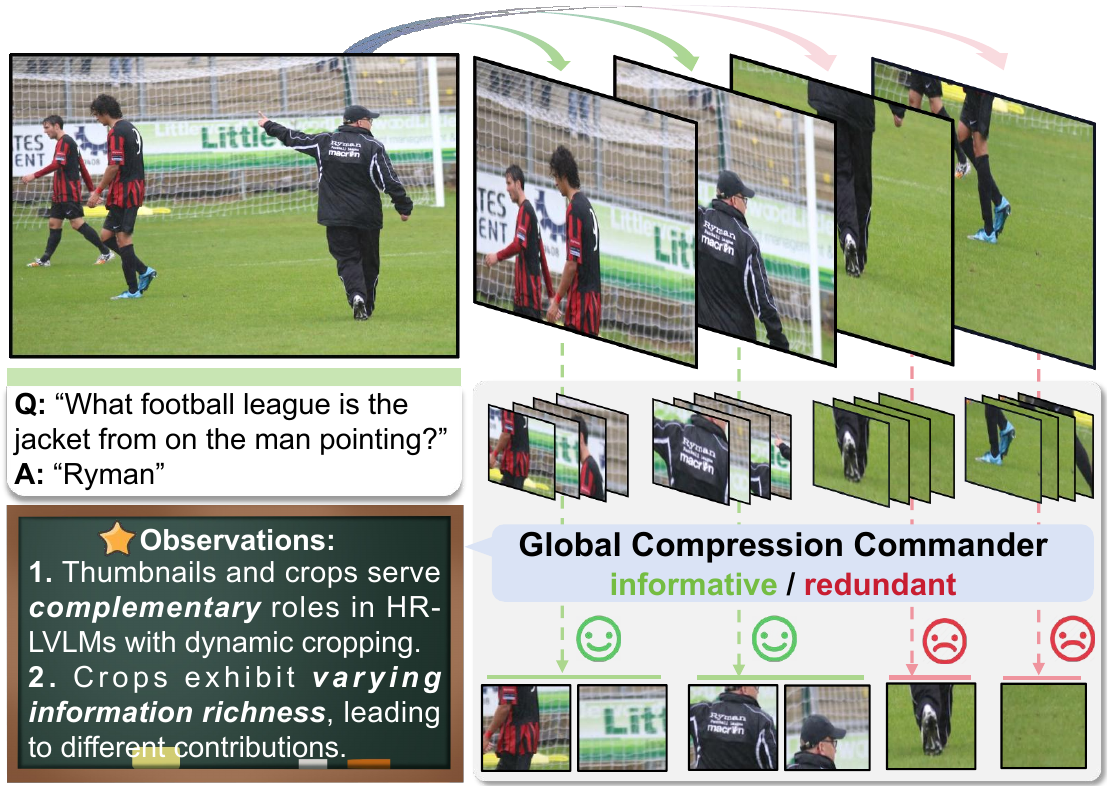}
  \caption{\textbf{Design philosophy of ``global-to-local'' guided token compression.} GlobalCom$^2$ evaluates the \textit{information richness} of local crops from a global perspective to preserve informative regions while removing redundant ones.
  }
  \label{fig:design_philosophy}
\end{figure}

To enhance the efficiency of LVLMs, recent efforts have increasingly adopted token compression approaches~\cite{Cha:Honeybee,Shang:LLaVA-PruMerge,liu2025video,liu2025shifting,liu2025mixkv,wen2025efficient}, which reduce visual tokens while preserving essential information. These architecture-agnostic methods achieve optimal efficiency-accuracy trade-offs and have proven effective for LVLM acceleration. However, these methods were primarily designed for traditional \textit{single-view} architectures (the entire image in Figure~\ref{fig:design_philosophy} top). While the \textit{multi-view} dynamic cropping approach enables more fine-grained visual understanding, it further increases the number of visual tokens and computational overhead.

\begin{figure}[t]
    \centering
    \includegraphics[width=\linewidth]{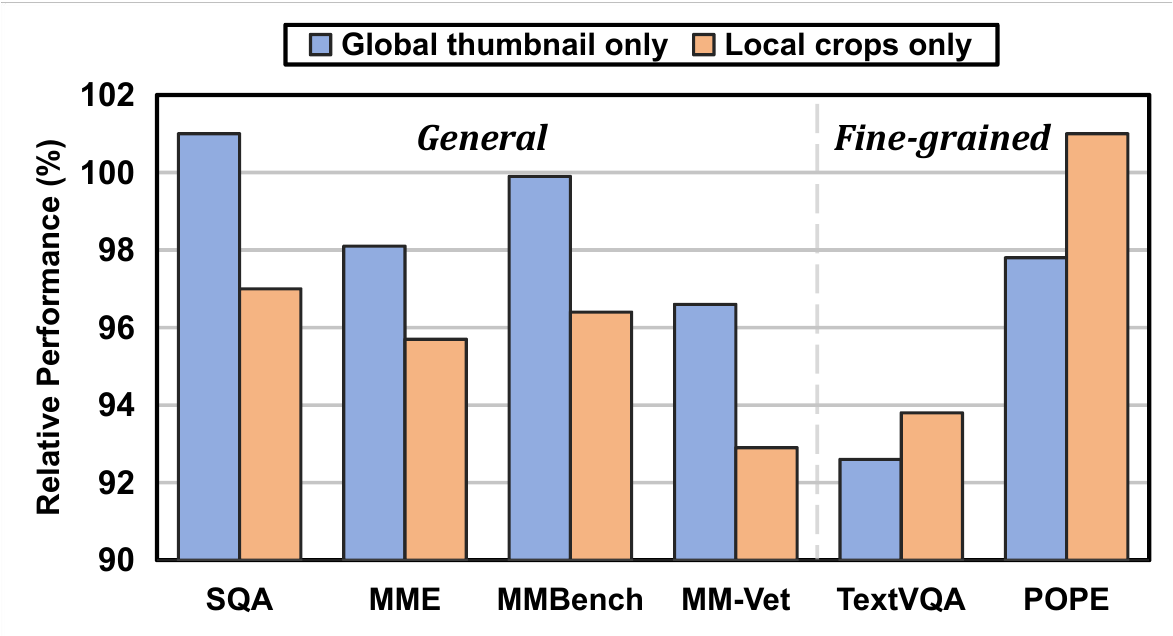}
    \caption{\textbf{Complementary roles of global thumbnail and local crops in HR-LVLMs with dynamic cropping.} Performance (\%) denotes relative scores of LLaVA-NeXT-7B.}
\label{fig:thumbnail and crops}
\end{figure}

\begin{figure}[t]
    \centering
    \includegraphics[width=\linewidth]{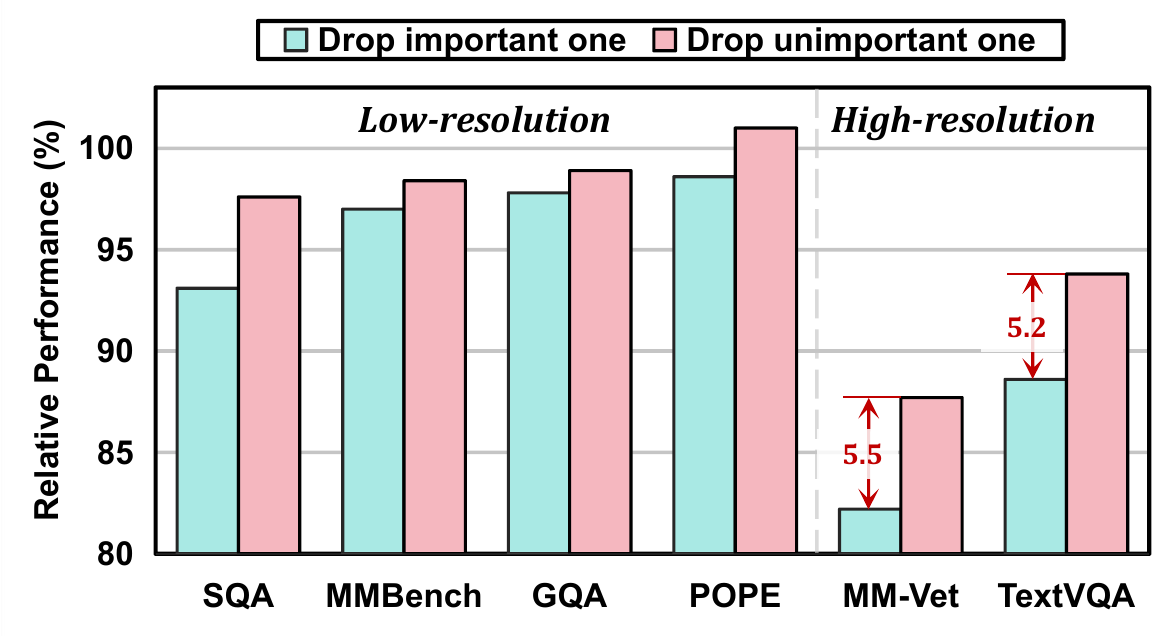}
    \caption{\textbf{Varying contributions of local crops.} Importance is quantified by the accumulated attention scores between thumbnail patches and \texttt{[CLS]} token within each crop.
    }
\label{fig:different crops}
\end{figure}

\begin{figure}[t]
    \centering
    \includegraphics[width=\linewidth]{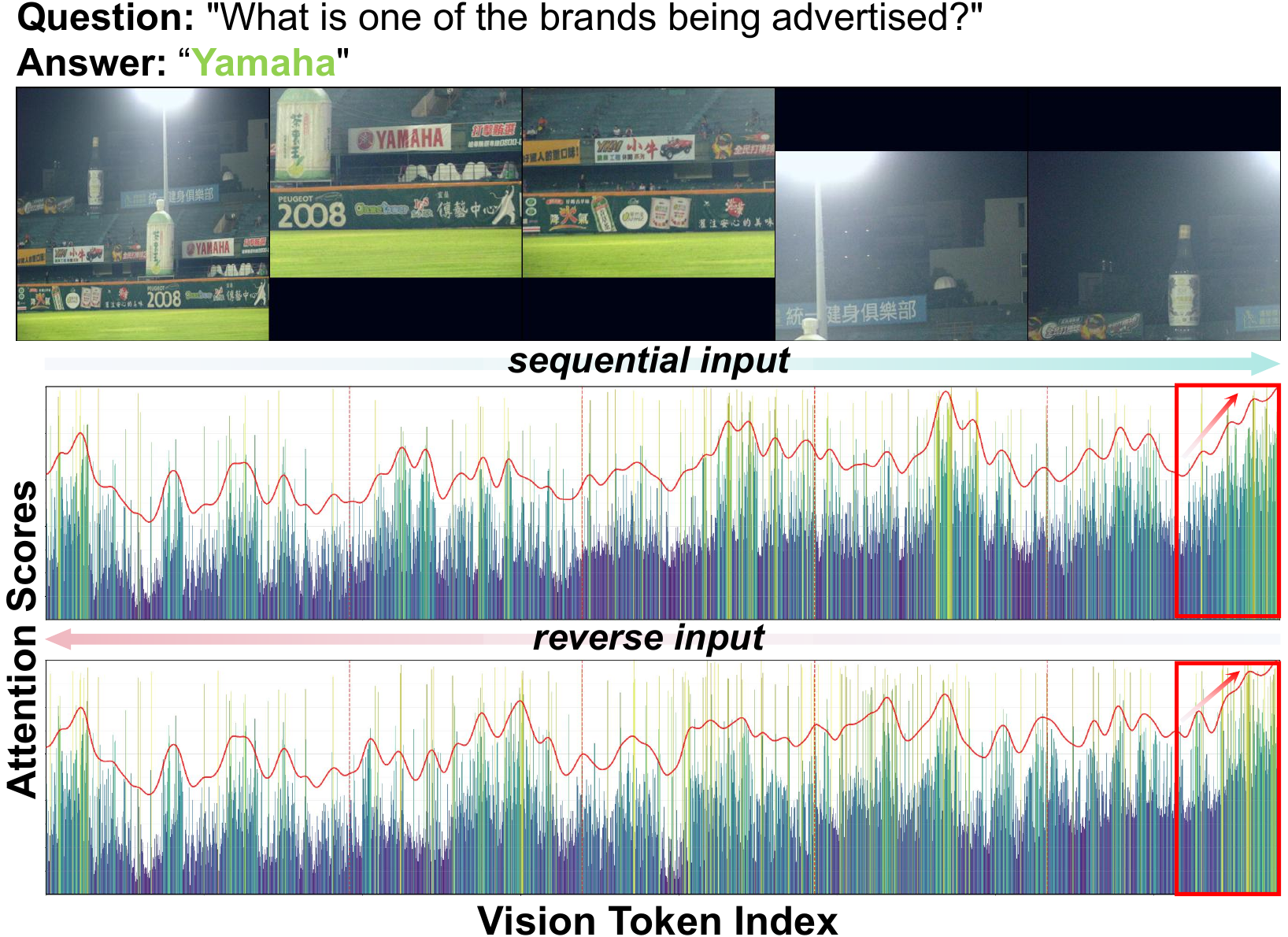}
    \caption{\textbf{Content-agnostic positional bias.} LLM attention-guided methods (\textit{e.g.}, FastV) assign higher scores (bars) to later tokens, regardless of their content or input order (second row: sequential crops; third row: reversed crops).
    }
\label{fig:positional_bias}
\end{figure}

However, directly applying existing token compression methods to HR-LVLMs overlooks \textbf{\textit{three critical issues}}: \textbf{(i) Global context neglect:} Current methods disregard the crucial role of global thumbnails in extracting holistic context and guiding visual understanding (Figure~\ref{fig:thumbnail and crops}). \textbf{(ii) Information richness disparity:} They fail to account for varying information density across different crops (Figure~\ref{fig:different crops}), resulting in performance gaps exceeding \textbf{5\%} on high-resolution tasks when comparing the removal of most versus least informative crops. \textbf{(iii) Content-agnostic positional bias:} Inner-LLM question-aware compression methods~\cite{Chen:FastV,Xing:PyramidDrop,Zhang:SparseVLM} systematically allocate more tokens to later-positioned crops irrespective of their actual informative content (Figure~\ref{fig:positional_bias}), inducing severe multi-modal hallucinations under extreme compression (Table~\ref{tab:main_results_7B}). These three oversights collectively result in \textbf{over-compression} of information-rich regions and disruption of the visual-semantic hierarchy (Figure~\ref{fig:compression_performance}).

To bridge this critical gap, we first conduct systematic analysis of dynamic cropping in Section~\ref{sec:Analysis}, identifying \textbf{\textit{two key observations}}: \textbf{\ding{182}} Thumbnail and crop tokens serve complementary roles - thumbnails capture holistic context while crops provide fine-grained details, enabling global importance evaluation. \textbf{\ding{183}} Through global context, tokens from different crops exhibit varying informativeness, requiring differentiated compression to minimize information loss.

Building on these observations, we propose ``\textbf{Global} \textbf{Com}pression \textbf{Com}mander'' (\textbf{GlobalCom$^2$}), which follows a ``\textbf{\textit{global-to-local}}'' guided token compression philosophy tailored for dynamic cropping-based HR-LVLMs. In Figure~\ref{fig:design_philosophy}, GlobalCom$^2$ leverages holistic information from thumbnails to evaluate each crop's information richness, adaptively adjusting compression intensity. It performs token compression through comprehensive evaluation from both global and local perspectives, achieving differentiated compression across regions while preserving significant information. This approach can be integrated with existing question-aware methods. 
Notably, integrating this ``global-to-local'' design with FastV~\cite{Chen:FastV} and SparseVLM~\cite{Zhang:SparseVLM} achieves significant improvements of \textbf{5.3\%} and \textbf{5.2\%} across benchmarks while alleviating multi-modal hallucination caused by positional bias.

To summarize, our main contributions are three-fold: 
\begin{itemize}

    \item \textbf{Systematic Dynamic Cropping Analysis:} We empirically analyze the hierarchical nature of dynamic cropping and thoroughly investigate existing token compression methods for HR-LVLMs, identifying fundamental causes of critical information over-compression.
        
    \item \textbf{Global-to-Local Compression Philosophy:} We propose GlobalCom$^2$, a training-free framework that adaptively adjusts compression intensity based on crop information richness assessment and preserves semantically important tokens across the entire visual hierarchy.
            
    \item \textbf{Superior Performance-Efficiency Trade-offs:} Extensive experiments on multiple HR-LVLMs with dynamic cropping demonstrate exceptional balance of GlobalCom$^2$, maintaining over 90\% performance with 90\% token reduction, while achieving substantial memory savings and throughput improvements.

\end{itemize}

\section{Related Work}
\label{sec:related work}


\noindent \textbf{High-resolution LVLMs.}
LVLMs integrate vision encoders and LLM decoders via projectors for feature alignment~\cite{liu2023llava,Dai:InstructBLIP}. Early LVLMs~\cite{Liu:LLaVA-1.5,Bai:Qwen-VL} resize images to fixed resolutions, causing shape distortion and detail loss. To address this, high-resolution LVLMs have emerged in three categories: \textbf{(i)} Hybrid resolution methods using dual visual encoders~\cite{Li:Mini-Gemini,Luo:LLaVA-HR}. \textbf{(ii)} Native resolution methods with NaViT-style encoders~\cite{Wang:Qwen2-VL,guo2025seed1}. \textbf{(iii)} Dynamic cropping methods~\cite{li2024llava-ov,zhu2025internvl3}, which split images into regions for single-encoder processing. Dynamic cropping has gained widespread adoption~\cite{lu2025bluelm,chen2025eagle2.5} due to its vision encoding efficiency. However, increased visual tokens introduce computational challenges in inference speed and memory usage. Our work focuses on improving the efficiency of dynamic cropping-based HR-LVLMs.


\noindent \textbf{Token Compression for LVLMs.}
Token compression, which aims to reduce the sequence length of tokens for computation efficiency, has been widely adopted for model acceleration~\cite{Rao:DynamicViT,Liang:EViT,Bolya:ToMe}.
For LVLMs, recent works have focused on training-free token compression for LVLM acceleration at two stages: \textbf{(i)} Vision Encoding stage~\cite{Shang:LLaVA-PruMerge,Yang2024:Visionzip}: FasterVLM~\cite{Zhang:FasterVLM} leverages \texttt{[CLS]} attention scores to prune visual tokens. \textbf{(ii)} LLM Pre-filling stage~\cite{Xing:PyramidDrop,Ye:PruneFit,wen2025dart}: FastV~\cite{Chen:FastV} prunes tokens based on LLM self-attention, while SparseVLM~\cite{Zhang:SparseVLM} uses cross-modal attention scores.
Some methods combine both stages~\cite{Liu2024:MUSTDrop,Han2024:FiCoCo}.
However, these methods treat all visual tokens equally in a ``flat'' token space, ignoring HR-LVLMs' hierarchical structure where thumbnails and crops serve distinct roles. 

Our work is the first to systematically quantify this oversight's severe consequences, revealing critical failure modes including uniform over-compression and positional bias. 
We address this by proposing a ``global-to-local'' compression framework that leverages hierarchical structure, representing a shift toward structure-aware token compression.

\section{Analysis of Dynamic Cropping}
\label{sec:Analysis}

\subsection{Preliminary}
\label{subsec:Preliminary}

We take LLaVA-NeXT, a widely-adopted HR-LVLM with dynamic cropping, as an example to conduct our analysis.

\noindent \textbf{Model Architecture.} Given input image and text, LLaVA-NeXT generates responses through: \textbf{(i)} Vision Encoding: ViT~\cite{Radford:CLIP} converts pixels to embeddings via MLP projector. \textbf{(ii)} LLM Decoding: LLM processes concatenated tokens, generating responses auto-regressively.

\noindent \textbf{Pre-processing.} LLaVA-NeXT uses grid templates: $\{2 \times 2, 1 \times \{2,3,4\}, \{2,3,4\} \times 1\}$. Images of size $W \times H$ are scaled to $(336 \times a) \times (336 \times b)$, yielding $n=a \times b$ local crops $\mathbf{X}^{L}$ and a thumbnail $\mathbf{X}^{G}$. Total token length is $(1+n) \times N$.



\noindent \textbf{Post-processing.} LLaVA-NeXT removes padding tokens and adds boundary tokens to mark image regions, preserving aspect ratios while enhancing efficiency.




    




\begin{figure*}[t]
    \centering
    \includegraphics[width=\textwidth]{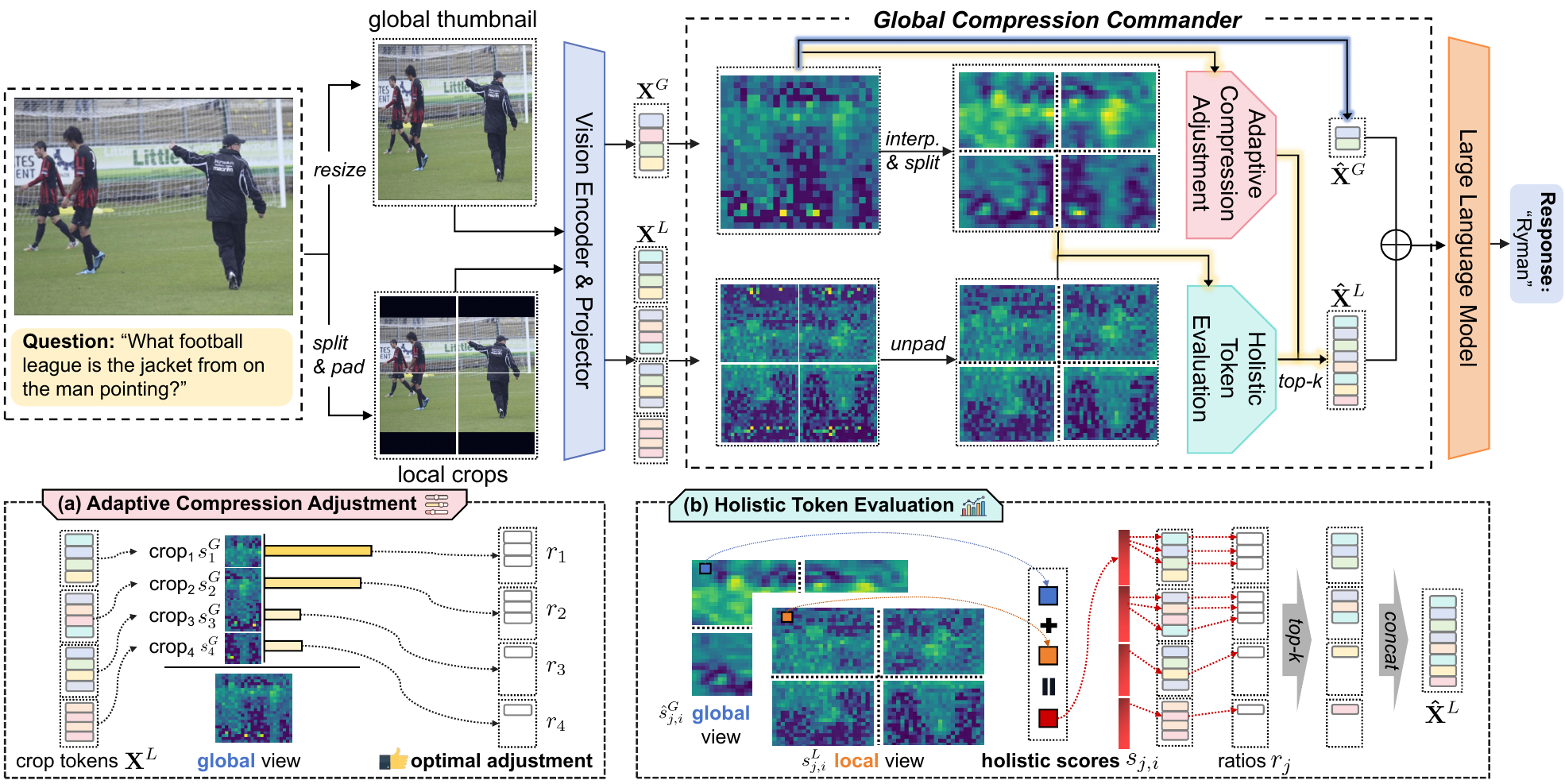}
    \caption{\textbf{Overall framework.} GlobalCom$^2$ guides token compression for HR-LVLMs through: 1) compressing thumbnail tokens (blue path), and 2) compressing crop tokens (yellow paths) by (a) adaptively adjusting compression intensity based on global visual richness, and (b) performing compression according to token informativeness from global and local perspectives.
    }
\label{fig:overview}
\end{figure*}

\subsection{Stepping into Dynamic Cropping}
\label{subsec:Analysis}

HR-LVLMs process high-resolution images using dynamic cropping with thumbnails and crops. We analyze their characteristics to guide token compression design.

\noindent \textbf{(I) Functions of Thumbnail and Crops:} We begin with an exploratory experiment to investigate how thumbnail and crops contribute to image understanding in HR-LVLMs by using them separately as input to LLaVA-NeXT-7B.

Figure~\ref{fig:thumbnail and crops} demonstrates that using global thumbnail alone yields superior performance on general visual perception benchmarks (\textit{e.g.}, SQA~\cite{Lu:ScienceQA} and MMBench~\cite{Liu:MMBench}). This advantage stems from global thumbnails providing holistic visual information through complete ViT encoding. In contrast, using only local crops, where ViT independently encodes each crop, shows inferior performance on these general visual perception tasks. However, local crops excel in fine-grained perception tasks like TextVQA~\cite{Singh:TextVQA} (VQA\textsuperscript{T}, testing text-centric visual understanding) and POPE~\cite{Li:POPE} (testing model hallucination) by providing detailed visual features. Therefore, we identify that:

\noindent \textbf{Observation \ding{182}:} Global thumbnails and local crops serve complementary functions in HR-LVLMs with dynamic cropping: the former acts as a ``comprehensive visual extractor'' for holistic representations, while the latter serves as a ``detailed visual capturer'' for fine-grained representations.

\noindent \textbf{(II) Information Richness in Different Crops:} Given that dynamic cropping leads to distinct visual representations across crops, we investigate the variance through both qualitative and quantitative perspectives.

Through visual inspection of the input image in Figure~\ref{fig:overview}, we observe that the upper crops contain rich visual content (\textit{e.g.}, football players), while the lower crops mainly show redundant grass areas. To quantitatively validate this observation, we analyze the attention scores between \texttt{[CLS]} and patch tokens, which have been shown to effectively indicate token importance in prior works~\cite{Liang:EViT,Han2024:FiCoCo}. As visualized in Figure~\ref{fig:overview}, the attention distribution from CLIP-ViT's last layer shows significantly higher values in the upper regions, confirming that upper crops contain more semantic information from a global perspective.

Building upon this qualitative analysis, we conduct quantitative studies to validate our observations. Since attention scores with the \texttt{[CLS]} token effectively measure token-level semantic richness, we leverage these scores to assess crop importance by by computing the sum of partitioned attention scores within each crop region. To examine crop contributions in HR-LVLMs, we conduct experiments on LLaVA-NeXT-7B by selectively dropping local crops (\textit{i.e.}, dropping the most/least important one). As shown in Figure~\ref{fig:different crops}, dropping the most versus least important crops leads to significant performance gaps across visual understanding tasks, with an average drop of \textbf{2.4\%} across six benchmarks and particularly notable in VQA\textsuperscript{T} (\textbf{5.2\%} gap). Based on the above analysis, we identify that:

\noindent \textbf{Observation \ding{183}:} Local crops exhibit varying information richness in global context, leading to different contributions to the overall visual understanding of HR-LVLMs with dynamic cropping, with visually informative crops being particularly crucial for capturing fine-grained local details.

\section{Global Compression Commander}
\label{sec:Methodology}

HR-LVLMs with dynamic cropping increase token length (\textbf{5×} in LLaVA-NeXT~\cite{Liu:LLaVA-NeXT}, \textbf{10×} in LLaVA-OV~\cite{li2024llava-ov}), making self-attention's quadratic complexity a bottleneck. Inspired by human vision's process of grasping scene gist before focusing on details, we propose ``\textbf{Global} \textbf{Com}pression \textbf{Com}mander'' (\textbf{GlobalCom$^2$}) which implements a ``global-to-local'' compression strategy.

In Figure~\ref{fig:overview}, for thumbnail token compression, we identify tokens with essential holistic information. Given the \texttt{[CLS]} token's effectiveness as global image representation~\cite{Liang:EViT}, GlobalCom$^2$ uses the last ViT layer's attention map to compute attention between each token and \texttt{[CLS]} (blue path in Figure~\ref{fig:overview}). For the 1D token sequence $\mathbf{X}^{G}$ of length $N$, the importance score ${s}_i^{G}$ for the $i$-th token is:
{\setlength\abovedisplayskip{2mm}
\setlength\belowdisplayskip{2mm}
\begin{equation}
\mathbf{s}_i^{G} = \frac{ \exp{( \mathbf{q}^{\texttt{CLS}} {\mathbf{K}}_i^\top / \sqrt{D} )} }{ \sum^N_{i=1} \exp{( \mathbf{q}^{\texttt{CLS}} {\mathbf{K}}_i^\top / \sqrt{D} )} },
\label{eq:CLS computation}
\end{equation}
}
where $\mathbf{q}^{\texttt{CLS}}$ and $\mathbf{K} \in \mathbb{R}^{N \times D}$ are query of $\texttt{[CLS]}$ and keys of $\mathbf{X}^{G}$. Given a preset retention ratio $R$ (\%), GlobalCom$^2$ preserves the top-$k$ ($k=R \times N$) tokens ranked by ${s}^{G}$:
{\setlength\abovedisplayskip{2mm}
\setlength\belowdisplayskip{2mm}
\begin{equation}
    \mathbf{X}^G \rightarrow \mathbf{\hat{X}}^G = \text{TopK}(\mathbf{X}^G, s^G, R \times N).
\label{eq:global compression}
\end{equation}
}

While thumbnail compression focuses on preserving holistic context, crop compression faces more complex challenges due to varying information densities across different crops. Following observation \textbf{\ding{183}}, semantically rich crops should preserve more tokens for crucial details, while less informative ones can be compressed more aggressively.

GlobalCom$^2$ leverages the comprehensive visual knowledge from thumbnails to guide crop compression through a decoupled two-stage process: \textbf{(a)} Adaptive Compression Adjustment, which dynamically adapts the compression intensity for each crop based on its information richness, and \textbf{(b)} Holistic Token Evaluation, which assesses the informativeness of tokens from both local and global views.

\subsection{Adaptive Compression Adjustment}
\label{subsec:Adaptive Compression Adjustment}

\newcommand{\downtiny}[1]{{\!\tiny{#1}}}

\begin{table*}[!t]
\centering
\tablestyle{5pt}{1.0}
\setlength\tabcolsep{8.1pt}
\def\w{20pt} 
\scalebox{1.08}{
    \begin{tabular}{lccccccccc}
    \textbf{Method} & \textbf{GQA} & \textbf{VizWiz} & \textbf{SQA} & \textbf{MMB} & \textbf{POPE} & \textbf{VQA\textsuperscript{T}} & \textbf{MME} & \textbf{MM-Vet} & \textbf{Average} \\
    \shline
    \multicolumn{10}{l}{\textit{Upper Bound, 2880 Tokens}} \\
    LLaVA-NeXT-7B & 64.2 & 57.6 & 70.1 & 67.4 & 86.5 & 64.9 & 1519.0 & 43.9 & 100.0\% \\
    \hline
    \multicolumn{10}{l}{\textit{Ratio=50\%, Retain up to 1440 Tokens}} \\
        FastV\tiny\texttt{(ECCV24)}  & 61.8 & 54.9 & 69.0 & 67.4 & 85.5 & 59.6 & 1490.3 & 37.6 & 95.5\% \\ 
        PDrop\tiny\texttt{(CVPR25)} & 63.7 & \textbf{57.9} & \textbf{69.2} & \textbf{67.7} & 87.9 & 61.6 & 1499.6 & 37.5 & 97.4\% \\
        SparseVLM\tiny\texttt{(ICML25)} & 63.7 & 57.2 & 68.3 & 67.6 & 87.9 & 60.5 & 1507.2 & 36.8 & 96.8\% \\
        FasterVLM\tiny\texttt{(2024.12)} & 63.4 & 56.4 & 69.1 & 67.4 & 87.7 & 58.9 & 1533.3 & 39.6 & 97.3\% \\
     \textbf{GlobalCom$^2$} & \textbf{63.9} & 56.5 & 68.5 & 67.6 & \textbf{88.1} & \textbf{62.3} & \textbf{1552.9} & \textbf{40.4} & \textbf{98.5\%} \\
    \hline
    \multicolumn{10}{l}{\textit{Ratio=25\%, Retain up to 720 Tokens}} \\
        FastV\tiny\texttt{(ECCV24)} & 60.4 & 54.2 & \textbf{68.8} & 65.6 & 83.1 & 58.4 & 1477.3 & 35.4 & 93.4\% \\ 
        PDrop\tiny\texttt{(CVPR25)} & 60.3 & \textbf{56.8} & 68.5 & 65.6 & 85.5 & 59.8 & 1473.7 & 31.1 & 93.3\% \\ 
        SparseVLM\tiny\texttt{(ICML25)} & 59.9 & 56.0 & 67.5 & 65.6 & 85.0 & 58.3 & 1465.9 & 38.5 & 94.6\% \\ 
        FasterVLM\tiny\texttt{(2024.12)} & 61.3 & 55.4 & 67.1 & \textbf{66.0} & 87.2 & 58.8 & 1454.6 & 37.8 & 94.8\% \\
        \textbf{GlobalCom$^2$} & \textbf{61.5} & 55.7 & 68.1 & 65.9 & \textbf{87.6} & \textbf{60.9} & \textbf{1493.5} & \textbf{40.7} & \textbf{96.7\%} \\
    \hline
    \multicolumn{10}{l}{\textit{Ratio=10\%, Retain up to 288 Tokens}} \\
        FastV\tiny\texttt{(ECCV24)} & 55.9 & 53.1 & 68.1 & 61.6 & 71.7 & 55.7 & 1282.9 & 27.2 & 85.4\% \\
        PDrop\tiny\texttt{(CVPR25)} & 54.5 & 54.4 & 67.7 & 59.0 & 77.6 & 54.4 & 1262.1 & 24.0 & 84.3\%  \\
        SparseVLM\tiny\texttt{(ICML25)} & 56.3 & 52.1 & 68.5 & 60.0 & 80.1 & 53.9 & 1334.2 & 26.5 & 86.1\% \\ 
        PruMerge\tiny\texttt{(ICCV25)} & 53.6 & 54.0 & 66.4 & 61.3 & 60.8 & 50.6 & 1149.3 & 25.5 & 80.6\% \\ 
        FasterVLM\tiny\texttt{(2024.12)} & 56.9 & 52.6 & 66.5 & 61.6 & 83.6 & 56.5 & 1359.2 & 35.0 & 89.9\% \\
        \textbf{GlobalCom$^2$} & \textbf{57.1} & \textbf{54.6} & \textbf{68.7} & \textbf{61.8} & \textbf{83.8} & \textbf{58.4} & \textbf{1365.5} & \textbf{36.4} & \textbf{91.6\%} \\
    \end{tabular}%
    }
    \caption{\textbf{Comparisons with LLaVA-NeXT across image understanding benchmarks.} VQA\textsuperscript{T} (TextVQA), MME, MM-Vet are high-resolution benchmarks. ``Average'' shows mean performance across benchmarks, with \textbf{best} results highlighted.}
  \label{tab:main_results_7B}%
\end{table*}%

As shown in bottom-left of Figure~\ref{fig:overview}, GlobalCom$^2$ analyzes each local crop's semantic contribution and applies adaptive token compression accordingly.

For differentiated compression, we compute each crop's \textit{information richness score} $s_j^G$ by accumulating patch-to-\texttt{[CLS]} attention scores in its corresponding global thumbnail region: $s_j^{G} = \sum_{i \in \text{crop}_j} s^{G}_i$. We then normalize scores with $\tilde{s}_j = (s_j^{G} - \max(s_j^{G}))/\tau$ ($\tau=10$) and compute relative importance weight $\sigma_j$ via softmax:
{\setlength\abovedisplayskip{2mm}
\setlength\belowdisplayskip{2mm}
\begin{equation}
    \sigma_j = \frac{\exp(\tilde{s}_j)}{\sum_{l=1}^{n} \exp(\tilde{s}_l) + \epsilon},
\label{eq:crop reletaive importance}
\end{equation}
}
where $\epsilon=10^{-8}$ prevents division by zero. The final retention ratio $r_j$ for each crop is adjusted from the preset ratio $R (\%)$ based on its global content importance:
{\setlength\abovedisplayskip{2mm}
\setlength\belowdisplayskip{2mm}
\begin{equation}
    r_j = R \times \left(1 + \sigma_j - \frac{1}{n}\right),
\label{eq:crop ratio allocation}
\end{equation}
}
where $\sigma_j - \frac{1}{n}$ is deviation from average importance, allocating more tokens to important content and fewer to unimportant, for adaptive content-aware compression.

Through this process, GlobalCom$^2$ allocates compression degrees (\textit{i.e.}, $\{r_1, r_2, r_3, r_4\}$ in Figure~\ref{fig:overview}) based on each crop's information richness from the global view.

\subsection{Holistic Token Evaluation}
\label{subsec:Holistic Token Evaluation}

After determining compression degrees for each crop, GlobalCom$^2$ evaluates token importance for preservation (Figure~\ref{fig:overview}). For each crop, attention between patch tokens and \texttt{[CLS]} yields \textit{local importance scores} $\{s_{j}^{L}\}_{j=1}^{n}$ from final-layer attention. These only capture within-crop importance, missing cross-region elements. Following observation \textbf{\ding{182}}, GlobalCom$^2$ incorporates global thumbnail context by reshaping 1D attention scores ${s}^{G}$ to 2D format and applying bilinear interpolation to match original dimensions, yielding $\{\hat{s}_{j}^{G}\}_{j=1}^{n}$ sub-maps per crop. The \textit{holistic score} $s_{j,i}$ for the $i$-th token in the $j$-th crop is:
{\setlength\abovedisplayskip{2mm}
\setlength\belowdisplayskip{2mm}
\begin{equation}
    s_{j,i} = \alpha \hat{s}_{j,i}^{G} + (1-\alpha) s_{j,i}^{L},
\label{eq:holistic token evalutation}
\end{equation}
}
where we empirically set $\alpha=0.5$ to give equal consideration to both information sources. The final compression:
{\setlength\abovedisplayskip{2mm}
\setlength\belowdisplayskip{2mm}
\begin{equation}
    \mathbf{X}^L_j \rightarrow \mathbf{\hat{X}}^L_j = \text{TopK}(\mathbf{X}^L_j, s_j, r_j \times N).
\end{equation}
}
This holistic evaluation identifies globally significant tokens while preserving local details.

\subsection{GlobalCom$^2$ without \texttt{[CLS]} Token}
\label{subsec:non-cls}

For HR-LVLMs without \texttt{[CLS]} token (\textit{e.g.}, LLaVA-OneVision~\cite{li2024llava-ov} with SigLIP~\cite{ZhaiM0B23:SigLIP}), we propose an alternative token informativeness measure for GlobalCom$^2$. Specifically, given a sequence of vision tokens $\mathbf{X} \in \mathbb{R}^{N \times d}$ after vision encoding, we first compute a global mean vector $\mathbf{g} \in \mathbb{R}^d$ through global average pooling over all tokens, and then calculate the cosine similarity between each patch token $\mathbf{x}_i$ and $\mathbf{g}$:

\begin{equation}
    c_i = \cos(\mathbf{x}_i, \mathbf{g}) = \frac{\mathbf{x}_i \cdot \mathbf{g}}{\|\mathbf{x}_i\| \|\mathbf{g}\|},
\end{equation}
The informativeness score $s_i = -c_i$ is negatively correlated with the similarity, reflecting that tokens with greater difference from $\mathbf{g}$ carry more unique information. Specifically, tokens exhibiting low similarity to $\mathbf{g}$ represent distinctive and irreplaceable visual elements, while highly similar tokens typically correspond to redundant or common patterns. This scoring mechanism serves as an effective alternative to \texttt{[CLS]}-based scoring in Equation~\ref{eq:global compression}-\ref{eq:holistic token evalutation} for models without \texttt{[CLS]} token. It is adopted to evaluate both crop-level information richness and token-level importance in LLaVA-OneVision. We conduct comprehensive quantitative and qualitative analyses in Appendix to explore how to measure token informativeness without \texttt{[CLS]} token.

\section{Experiments}
\label{sec:Experiments}

\subsection{Experimental Setting}
\label{subsec:Setting}

We evaluate on LLaVA-NeXT~\cite{Liu:LLaVA-NeXT} and LLaVA-OneVision~\cite{li2024llava-ov} for evaluation. We compare with FastV~\cite{Chen:FastV}, SparseVLM~\cite{Zhang:SparseVLM}, PDrop~\cite{Xing:PyramidDrop}, PruMerge~\cite{Shang:LLaVA-PruMerge}, FasterVLM~\cite{Zhang:FasterVLM} at different retention ratios $R$. For fair comparison with multi-stage methods, we use ``equivalent token count'' reflecting the average percentage of vision tokens retained across all LLM layers.

\subsection{Main Results}
\label{subsec:Comparison}

\begin{figure}[!t]
  \centering
   \includegraphics[width=\linewidth]{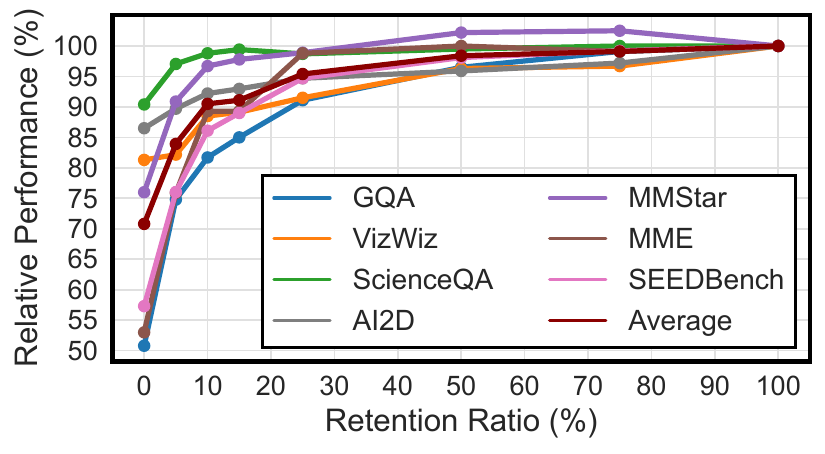}
   \caption{\textbf{Results of GlobalCom$^2$ on LLaVA-OneVision for image understanding.} GlobalCom$^2$ achieves 90.5\% average performance with only 10\% visial tokens.}
   \label{fig:llava-ov}
\end{figure}

\begin{figure*}[!t]
  \centering
   \includegraphics[width=\linewidth]{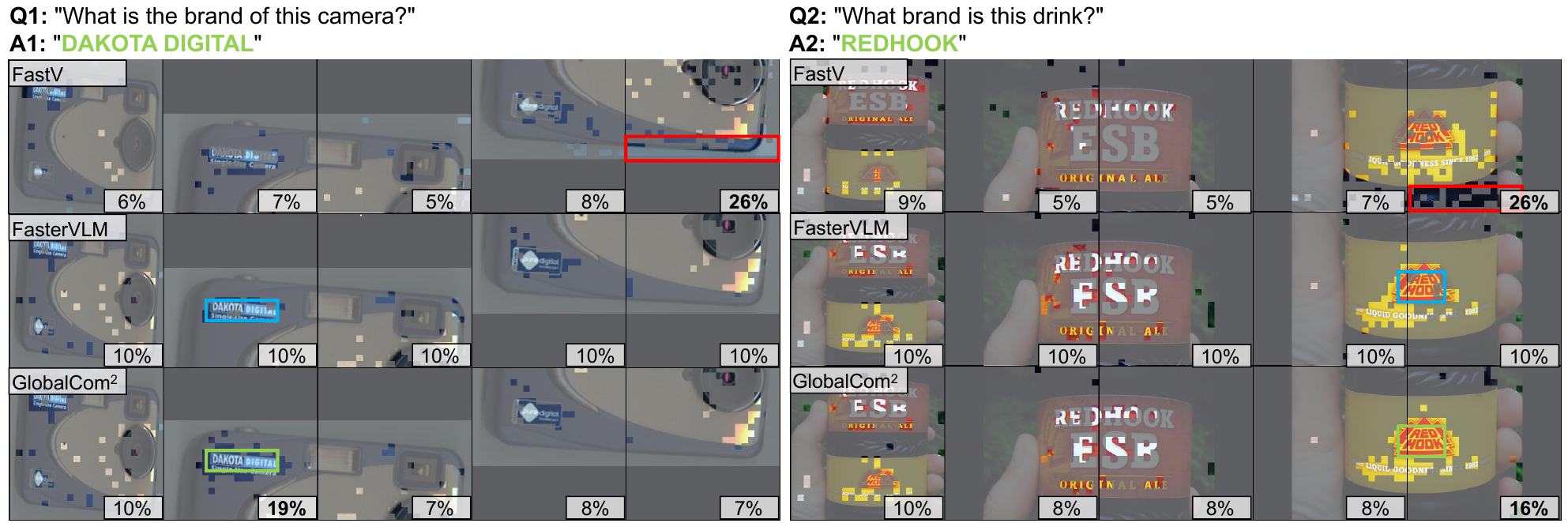}
    \caption{\textbf{Visualization of different token compression methods.} Gray masks indicate discarded tokens, where other methods exhibit significant \textit{over-compression} issues, while GlobalCom$^2$ preserves both global important and local detailed information.}   
   \label{fig:compression_performance}
\end{figure*}

\noindent \textbf{Results on LLaVA-NeXT.}
Table~\ref{tab:main_results_7B} compares GlobalCom$^2$ with existing methods with LLaVA-NeXT, which demonstrates its three key advantages: \textbf{(i) Superior performance:} GlobalCom$^2$ consistently outperforms all baselines, being the only method that maintains \textbf{>90\%} of the original performance across all retention ratios. \textbf{(ii) Extreme compression robustness:} While baseline methods suffer significant degradation at $R=10\%$, GlobalCom$^2$ maintains robust performance and achieves the best results across all benchmarks, outperforming the second-best method by \textbf{1.7\%} on average. Notably, FastV and PDrop demonstrate severe deterioration on POPE, exhibiting clear multi-modal hallucination due to positional bias from attention-guided token selection (Figure~\ref{fig:compression_performance}). \textbf{(iii) High-resolution excellence:} GlobalCom$^2$ demonstrates exceptional performance on high-resolution benchmarks (\textit{e.g.}, VQA\textsuperscript{T}, MME, MM-Vet). Our ``global-to-local'' guided compression preserves both global semantics and local details, outperforming baselines that suffer from over-compression.

\noindent \textbf{Results on LLaVA-OneVision.}
Figure~\ref{fig:llava-ov} further presents GlobalCom$^2$'s performance across benchmarks under varying retention ratios $R$ on LLaVA-OneVision. Generally, model performance correlates with $R$, with more aggressive compression leading to degradation. Vision-centric tasks (\textit{e.g.}, GQA, VizWiz, MME, SEED) show substantial drops with reduced visual tokens, while SQA maintains robust performance even with minimal tokens, suggesting language understanding dominates in scientific reasoning. Notably, GlobalCom$^2$ preserves \textbf{90.5\%} performance at $R=10\%$ while consuming only \textbf{35.4\%} of the original GPU memory, all without any training overhead, demonstrating its effective and efficient token compression.

\subsection{Ablation Study and Analysis}
\label{subsec:Ablation}

\noindent \textbf{Ablation on Adaptive Compression Adjustment.} 
Table~\ref{tab:ablation_1} compares four settings: (a) ``Uniform'' baseline with $R = 25\%$ across all crops, and three adaptive strategies: (b) ``$\mathbf{n_\text{top-\textit{k}}}$'' adjusting compression based on top-$k$ most informative tokens per crop ($k=25\% \times N$), (c) ``Softmax (max)'' applying softmax over maximum token importance score $s_j^G$ in each crop's thumbnail, and (d) ``Softmax (sum)'' (our choice) computing softmax over sum of token importance scores $s_j^G$ in each crop's thumbnail region.
All adaptive strategies outperform uniform compression, with ``Softmax (sum)'' achieving the best performance. While $\mathbf{n_\text{top-\textit{k}}}$ and ``Softmax (max)'' focus on strongest visual features per crop without considering global importance, ``Softmax (sum)'' adjusts compression based on each crop's overall importance to the entire image, preserving more semantic information and helping the LLM capture finer visual details.

  

\begin{table}[!t]
  \centering
  \setlength{\tabcolsep}{0.5pt}
  
    \begin{tabular}{lcccccc}
    \textbf{Method} & \textbf{SQA} & \textbf{POPE} & \textbf{VQA\textsuperscript{T}} & \textbf{MME} & \textbf{MM-Vet} & \textbf{Avg.} \\
    \shline
    \multicolumn{7}{l}{\textit{Upper Bound, 2880 Tokens}} \\
    Vanilla & 70.1 & 86.5 & 64.9 & 1519.0 & 43.9 & 100.0\% \\
    \hline
    \multicolumn{7}{l}{\textit{Ratio=25\%, Retain up to 720 Tokens}} \\
    Uniform & 67.1  & 87.2  & 60.1 & 1454.6 & 37.8  & 94.2\% \\
    \hline
    $\mathbf{n_\text{top-\textit{k}}}$ & 67.4 & 87.3  & 59.8 & 1471.5 & 35.7  & 94.5\% \\
    Softmax (max) & 67.3  & 87.2  & 60.3 & 1462.6 & 38.4  & 94.7\% \\
     \textbf{Softmax (sum)} & \textbf{67.6} & \textbf{87.4} & \textbf{60.6} & \textbf{1473.3} & \textbf{39.6} & \textbf{95.6\%} \\
    \end{tabular}%
  \caption{\textbf{Effects of different adjustment strategies.} ``Uniform'' performs no compression adjustment, while the other three strategies enable adaptive compression adjustment.}
  \label{tab:ablation_1}%
\end{table}%

  
\begin{table}[!t]
  \centering
  \setlength{\tabcolsep}{0.8pt} 
    \scalebox{0.95}{\begin{tabular}{lcccccc}
    \textbf{Method} & \textbf{SQA} & \textbf{POPE} & \textbf{VQA\textsuperscript{T}} & \textbf{MME} & \textbf{MM-Vet} & \textbf{Avg.} \\
    \shline
     \multicolumn{7}{l}{\textit{Upper Bound, 2880 Tokens}} \\
    Vanilla & 70.1 & 86.5 & 64.9 & 1519.0 & 43.9 & 100.0\% \\
    \hline
    \multicolumn{7}{l}{\textit{Ratio=25\%, Retain up to 720 Tokens}} \\
    Local only & 67.6  & 87.4  & 60.6  & 1473.3 & 39.6  & 95.6\% \\
    Global only & 67.9  & 86.4  & 60.2 & 1488.5 & 37.8  & 94.7\% \\
    \textbf{Global and Local} & \textbf{68.1} & \textbf{87.6} & \textbf{60.9} & \textbf{1493.5} & \textbf{40.7} & \textbf{96.7\%} \\
    \end{tabular}%
    }
  \caption{\textbf{Effects of different token evaluation metrics.} For the $i$-th token in crop $j$, ``Global'' and ``Local'' refer to importance scores $\hat{s}_{j,i}^{G}$ and $s_{j,i}^{L}$ in Equation~\eqref{eq:holistic token evalutation}, respectively.}
  \label{tab:ablation_2}%
\end{table}%

\noindent \textbf{Ablation on Holistic Token Evaluation.} Table~\ref{tab:ablation_2} compares different token evaluation strategies. While both strategies are effective, each has limitations: Local-only evaluation excels at fine-grained tasks (VQA\textsuperscript{T}, POPE) but underperforms on general perception benchmarks (MME, SQA) due to missing global context. Global-only evaluation maintains general perception but may overlook crucial local details. GlobalCom$^2$ achieves optimal performance by combining them, leveraging their complementary strengths.

\begin{figure}[t]
    \centering
    \includegraphics[width=\linewidth]{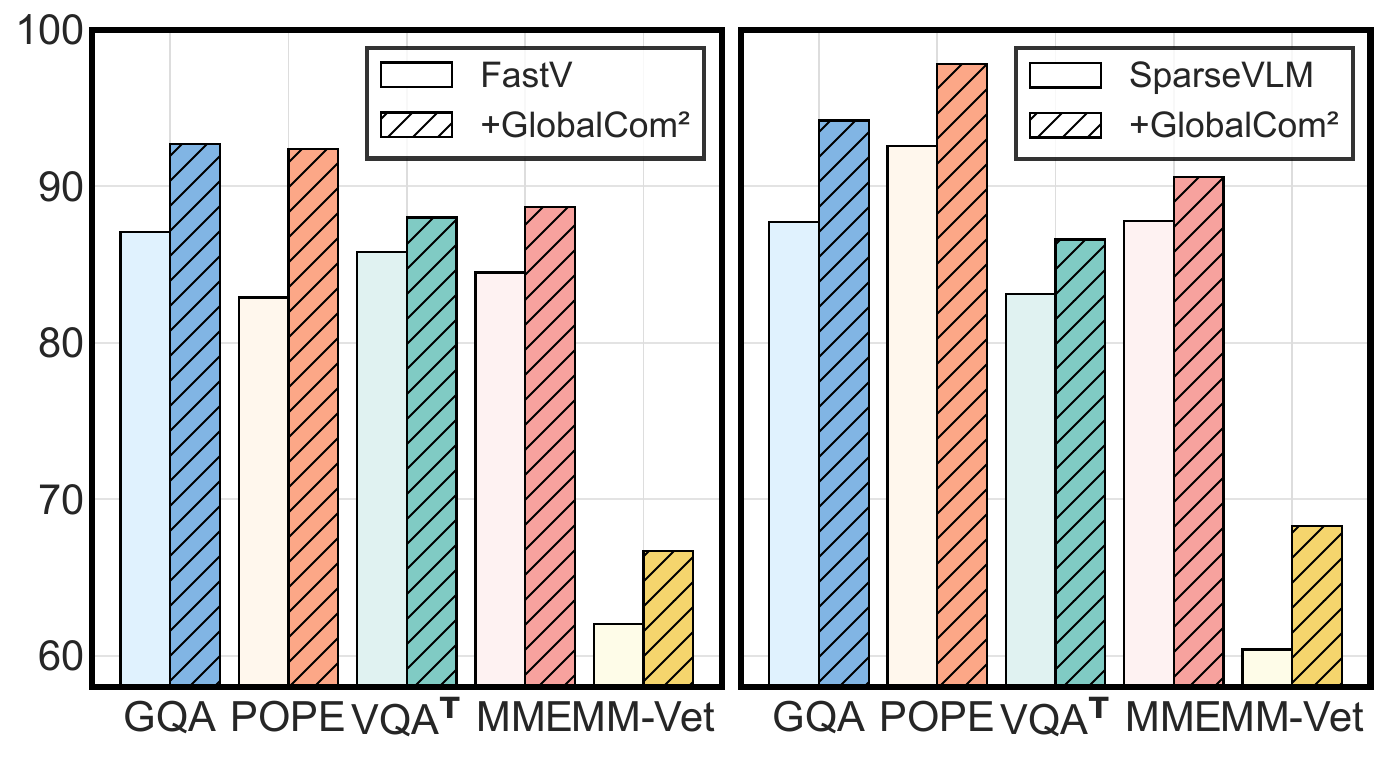}
    \caption{\textbf{Combination with question-aware methods.} ``+GlobalCom$^2$'' indicates the application of our Adaptive Compression Adjustment strategy under $R=10\%$.
    }
\label{fig:fastv_application}
\end{figure}

\noindent \textbf{Combination with Question-aware Methods.} Figure~\ref{fig:fastv_application} explores combining GlobalCom$^2$ with question-aware methods FastV and SparseVLM, enabling joint consideration of \textit{textual relevance and visual importance}. Using our Adaptive Compression Adjustment strategy, we assign optimal compression intensities (\textit{i.e.}, $r_j$ for the $j$-th crop) per crop based on its visual information richness within the global context before applying FastV/SparseVLM's token evaluation metrics. Under extreme compression ($R=10\%$), incorporating GlobalCom$^2$ yields average improvements of \textbf{5.3\%} and \textbf{5.2\%} for FastV and SparseVLM. Notably, GlobalCom$^2$ significantly boosts performance on POPE, with improvements of \textbf{8.2} for FastV and \textbf{4.5} for SparseVLM, confirming our strategy's effectiveness in \textit{mitigating positional bias} by preventing over-compression of important regions.

\subsection{Efficiency Analysis}
\label{subsec:efficiency}

Table~\ref{tab:efficiency comparisons} compares inference efficiency among SparseVLM, FasterVLM and GlobalCom$^2$. SparseVLM requires explicit attention scores in LLM and is naturally incompatible with FlashAttention~\cite{Dao2022:FlashAttention}, leading to higher memory costs~\cite{wen2025token}. Instead, FasterVLM and our GlobalCom$^2$ enable efficient computation before LLM decoding for efficient computation. As a plug-and-play solution, GlobalCom$^2$ achieves superior performance-efficiency trade-offs, maintaining \textbf{90\%} of the original performance while dramatically reducing peak memory usage by \textbf{40\%} and boosting inference throughput by \textbf{$1.8\times$}. 

\subsection{Compression Visualizations}
\label{subsec:Visualizations}

Figure~\ref{fig:compression_performance} compares compression processes of different methods at extreme compression setting of $R=10\%$ on VQA\textsuperscript{T}, revealing over-compression issues in baselines: \textbf{(i) Positional Bias:} FastV exhibits clear positional bias, allocating more tokens (>$3\times$) to later-positioned crops regardless of visual content. \textbf{(ii) Uniform Compression:} FasterVLM applies uniform compression across all crops, treating them as equally important. It fails to preserve critical information in some regions while retaining redundancy in others. In contrast, GlobalCom$^2$ globally assesses crop informativeness and adaptively adjusts compression ratios, preserving crucial details while eliminating redundancy.

\definecolor{greenrightcolor}{RGB}{0,144,81} 
\definecolor{redrightcolor}{RGB}{255,0,0} 
\definecolor{bluerightcolor}{RGB}{0,0,255} %

\newcommand{\downtinya}[1]{{\!\textcolor{greenrightcolor}{\tiny{#1}}}}
\newcommand{\uptiny}[1]{{\!\textcolor{redrightcolor}{\tiny{#1}}}}
\newcommand{\downtinyb}[1]{{\!\textcolor{bluerightcolor}{\tiny{#1}}}}

\newcommand{\cmark}{\ding{51}}%
\newcommand{\cmarkdark}{\ding{51}}%
\newcommand{\xmark}{\ding{55}}%

    

    

\begin{table}[!t]
  \centering
  \setlength{\tabcolsep}{0.2pt} 
    \scalebox{0.93}{\begin{tabular}{lcccc}
    \textbf{Method} & \textbf{TFLOPs↓} & \textbf{Memory↓} & \textbf{Throughput↑} & \textbf{Performance↑} \\
    \shline
     \multicolumn{3}{l}{\textit{Upper Bound, 2880 Tokens}} \\
    Vanilla & 41.7 & 23.0 & 3.8 & 100\% \\
    
    \hline
    \multicolumn{4}{l}{\textit{Ratio=10\%, Retain up to 288 Tokens}} \\

    SparseVLM & 5.4 \downtiny{(↓87.1\%)} & 24.2 \downtiny{(↑5.2\%)}  & 5.9 \downtiny{(1.6$\times$)} & 85.7\% \\

    FasterVLM & \textbf{3.8 \downtiny{(↓90.9\%)}} & \textbf{13.6 \downtiny{(↓40.1\%)}} & \textbf{6.7 \downtiny{(1.8$\times$)}} & 89.5\% \\
    
    \textbf{GlobalCom$^2$} & \textbf{3.8 \downtiny{(↓90.9\%)}} & 13.9 \downtiny{(↓40.0\%)} & \textbf{6.7 \downtiny{(1.8$\times$)}} & \textbf{90.8\%} \\
    \end{tabular}%
    }
  \caption{\textbf{Efficiency comparisons.} ``Memory'': peak GPU memory; ``Throughput'': POPE samples/second; ``Performance'': average score on eight multi-modal benchmarks.}
  \label{tab:efficiency comparisons}%
\end{table}%

\section{Conclusion}
\label{sec:Conclusion}

Token compression has achieved significant progress in accelerating LVLM inference. When applying existing methods to HR-LVLMs with dynamic cropping, these methods treat global thumbnails and local crops uniformly, overlooking their inherent characteristics. Through analyzing HR-LVLMs with dynamic cropping, we reveal distinct roles between thumbnails and crops, and observe varying information densities across crops. Based on these findings, we propose GlobalCom$^2$, a plug-and-play token compression framework that operates on a ``global-to-local'' guided principle, adaptively preserving informative regions while minimizing redundancy. Experiments show that GlobalCom$^2$ achieves superior performance and efficiency across benchmarks, significantly outperforming existing baselines.

\section*{Acknowledgments}
This work was supported in part by the Chengdu Science and Technology Program (No. 2025-YF12-00006-RC) , Police Integration Computing Key Laboratory of Sichuan Province (No. JWRH202502002), and the Open Fund of Key Laboratory of the Ministry of Education on Artificial Intelligence in Equipment (No. 2024-AAIE-KF04-03).

\bibliography{ref}

\clearpage
\appendix

\section*{Appendix}


In the appendix, we provide theoretical FLOPs calculation, more discussions about content-agnostic positional bias, GlobalCom$^2$ without \texttt{[CLS]} token, detailed benchmarks introduction and implementation details, more additional experiments and analysis, and detailed algorithm.

\section{Theoretical Complexity Analysis}
\label{sec:FLOPs}

GlobalCom$^2$ compresses visual tokens for HR-LVLMs, thereby reducing their computational costs. Below, we analyze the theoretical computational complexity of HR-LVLMs in both the prefill stage and the decoding stage.

During the prefill stage, the FLOPs for a single transformer layer can be estimated using the formula $8 T d^2+4 T^2 d+6 T d m$. When applying a token retention ratio $R$, where the retained token count is defined as $\hat{T}=R \cdot T$, the corresponding theoretical FLOPs reduction ratio $\eta$ can be reformulated to account for this adjustment:

\begin{equation}
\begin{split}
\eta &= 1-\frac{8 \hat{T} d^2+4 \hat{T}^2 d+6 \hat{T} d m}{8 T d^2+4 T^2 d+6 T d m} \\
     &= 1-\frac{R(8 d+4 R T +6 m)}{8 d+4 T +6 m}
\end{split}
\end{equation}

In the decoding stage, the integration of a KV-Cache substantially enhances computational efficiency. This improvement is evidenced by the reduction in the complexity of attention computation to $\mathcal{O}(T)$. As a result, the formula for computing FLOPs is refined to $8 d^2+4 T d+6 T d m$. Given the current limitations of hardware, managing dynamic KV-Cache lengths effectively during the inference process presents significant challenges. Therefore, implementing pruning strategies prior to the decoder in large language models could facilitate a more efficient acceleration of the inference process.

\section{More Discussions about Content-agnostic Positional Bias.}
\label{sec:content-agnostic positional bias}

\begin{figure*}[t]
    \centering
    \includegraphics[width=\linewidth]{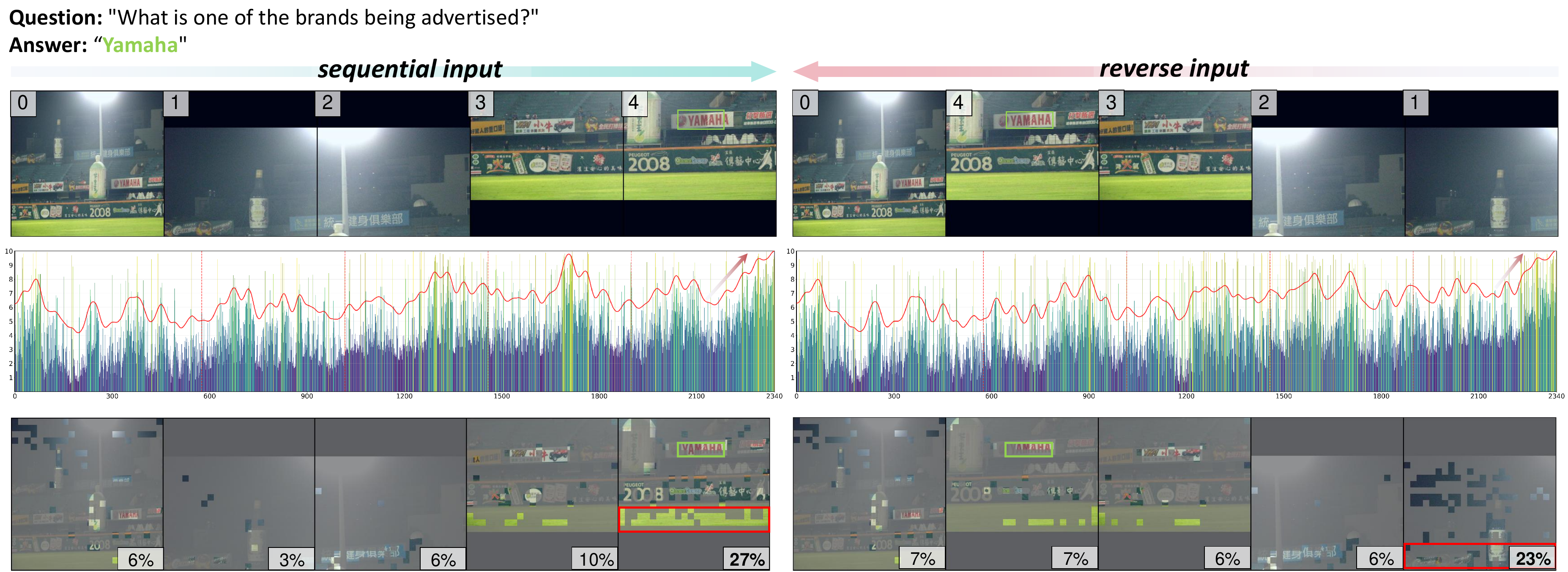}
    \vspace{-6mm}
    \caption{\textbf{Detailed analysis of content-agnostic positional bias.} LLM attention-guided methods (\textit{e.g.}, FastV~\cite{Chen:FastV}, PDrop~\cite{Xing:PyramidDrop}) assign disproportionately high scores (bars) to tokens in later positions, regardless of their actual semantic content or input order (second row: sequential crop input; third row: reverse crop input).
    }
    \vspace{-4mm}
\label{fig:detailed_positional_bias}
\end{figure*}

Figure~\ref{fig:detailed_positional_bias} reveals the inherent content-agnostic positional bias of the LLM attention-guided method FastV. Regardless of input order (sequential or reverse), FastV consistently assigns higher attention scores to later-positioned tokens when measuring visual token importance via LLM's second-layer attention. In extreme compression settings ($R=10\%$), this bias causes severe over-compression—preserving later tokens while discarding potentially more informative earlier ones, resulting in sub-optimal performance and significant multi-modal hallucinations (14.8-point drop on POPE).

\section{Discussions about LVLMs without \texttt{[CLS]}.}
\label{sec:more non-cls discussions}

Some Vision Encoders like SigLIP~\cite{ZhaiM0B23:SigLIP} do not own \texttt{[CLS]} tokens, rendering existing token compression methods that rely on \texttt{[CLS]} tokens (\textit{e.g.}, FasterVLM~\cite{Zhang:FasterVLM}) ineffective. Therefore, we explore alternative token importance evaluation approaches. This enables GlobalCom$^2$ to measure information content in each crop and adaptively allocate token budgets, preserving tokens with higher \textbf{information richness}.

\begin{figure*}[!t]
  \centering
   \includegraphics[width=\linewidth]{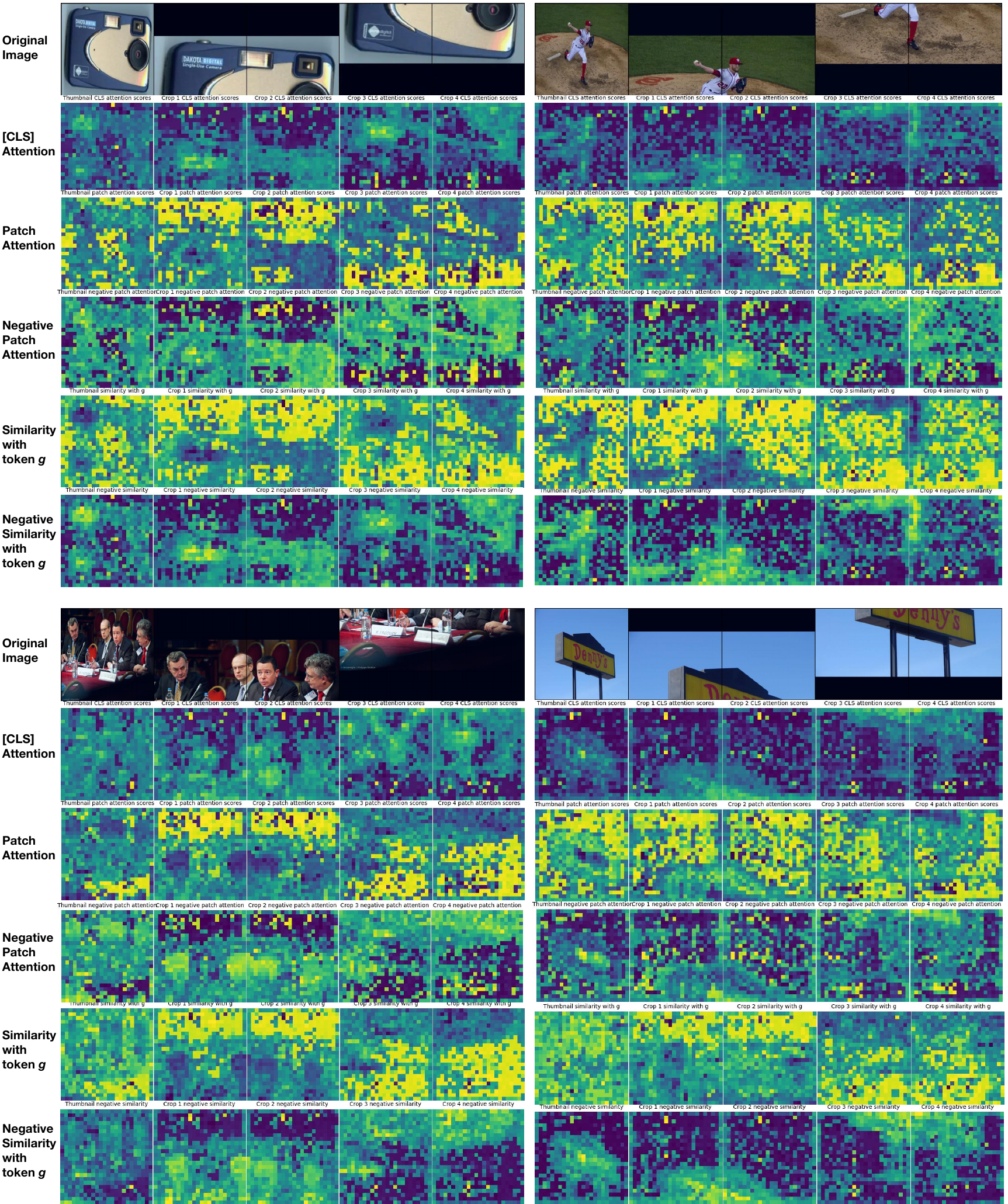}
   \vspace{-7mm}
    \caption{\textbf{Visualization of different token evalutaion scores.} }

   \label{fig:clip token evaluation scores}
   \vspace{-2mm}
\end{figure*}

As visualized in Figure~\ref{fig:clip token evaluation scores}, we first examine the attention patterns of CLIP-ViT's final layer \texttt{[CLS]} token as a baseline for evaluating alternative token importance assessment strategies. We explore two approaches to measure token information without relying on \texttt{[CLS]} token:

\textbf{(a) Patch Attention Analysis:} We calculate each patch token's importance by averaging its attention scores with all other patch tokens:
{\setlength\abovedisplayskip{2mm}
\setlength\belowdisplayskip{2mm}
\begin{equation}
    a_i = \frac{1}{N-1}\sum_{j\neq i} A_{i,j},
\end{equation}
}
where $A_{i,j}$ denotes the attention score from token $i$ to token $j$, and $N$ is the total number of patch tokens. Interestingly, as shown in the ``Patch Attention'' visualization in Figure~\ref{fig:clip token evaluation scores}, tokens with high informativeness exhibit \textbf{lower} average attention scores. This aligns with the intuition that tokens heavily attending to other tokens (high average attention scores) are more likely to be replaceable, as they primarily aggregate information from their neighbors. Conversely, tokens with low average attention scores tend to be more unique and irreplaceable, containing distinct information vital for image understanding. By negating these scores (shown as ``Negative Patch Attention''), we obtain patterns that relatively align with \texttt{[CLS]}-based assessment.

\textbf{(b) Global Mean Similarity:} We compute a global mean vector $\mathbf{g} \in \mathbb{R}^d$ through \textit{global average pooling} over all tokens, then calculate the cosine similarity between each patch token $\mathbf{x}_i$ and $\mathbf{g}$:
{\setlength\abovedisplayskip{2mm}
\setlength\belowdisplayskip{2mm}
\begin{equation}
    c_i = \text{sim}(\mathbf{x}_i, \mathbf{g}) = \frac{\mathbf{x}_i \cdot \mathbf{g}}{\|\mathbf{x}_i\| \|\mathbf{g}\|},
\end{equation}
}

Tokens with low similarity to $\mathbf{g}$ typically represent \textbf{distinctive} visual elements that \textit{deviate} from the average representation, suggesting their \textbf{irreplaceability} in capturing unique semantic information. Conversely, tokens highly similar to $\mathbf{g}$ often correspond to common or repetitive patterns, indicating their redundancy in visual representation. The ``Negative Similarity with token $\mathbf{g}$'' visualization aligns well with \texttt{[CLS]}-based assessment, highlighting semantically rich regions like the baseball player in the top-right example of Figure~\ref{fig:clip token evaluation scores}.

Based on these observations, we define two token importance measures:
{\setlength\abovedisplayskip{2mm}
\setlength\belowdisplayskip{2mm}
\begin{equation}
    s_i^{attn} = -a_i, \quad s_i^{sim} = -c_i,
\end{equation}
}
where $s_i^{attn}$ represents the ``Negative Patch Attention'' score and $s_i^{sim}$ denotes the ``Negative Similarity'' score. Both scores are designed to highlight informative tokens: $s_i^{attn}$ identifies tokens that maintain their distinctiveness by less attending to others, while $s_i^{sim}$ emphasizes tokens that deviate from the average representation $\mathbf{g}$. These measures serve as effective alternatives to \texttt{[CLS]}-based assessment (denoted as $s_i^{\texttt{[CLS]}}$) in our quantitative experiments.

\begin{table}[!t]
  \centering
  \setlength{\tabcolsep}{0.6pt} 
    \begin{tabular}{lccccccc}
    \textbf{Method} & \textbf{VQA\textsuperscript{T}} & \textbf{POPE} & \textbf{MME} & \textbf{MM-Vet} & \textbf{MMB} & \textbf{SQA} & \textbf{Avg.} \\
    \shline
     \multicolumn{7}{l}{\textit{Upper Bound, 2880 Tokens}} \\
    Vanilla & 64.9 & 86.5 & 1519.0 & 43.9 & 67.4 & 70.1 & 100.0\% \\
    \hline
    \multicolumn{7}{l}{\textit{Ratio=25\%, Retain up to 720 Tokens}} \\
    $s_i^{\texttt{[CLS]}}$ & 60.9  & 87.6  & 1493.5  & 39.5 & 65.9  & 68.1 & 96.4\% \\
    $s_i^{attn}$ & 58.7  & 85.7  & 1449.3 & 36.6 & 63.5  & 67.9 & 93.2\% \\
    $s_i^{sim}$ & 60.3 & 87.3 & 1485.8 & 39.7 & 65.0 & 67.6 & 95.8\% \\
    \end{tabular}%
  \vspace{-2mm}
  \caption{\textbf{Effects of different token evaluation metrics.}}
  \label{tab:non-cls}%
  \vspace{-2mm}
\end{table}%

Table~\ref{tab:non-cls} compares the effectiveness of $s_i^{\texttt{[CLS]}}$, $s_i^{attn}$, and $s_i^{sim}$ when used as metrics for both \textit{adaptive compression adjustment} and \textit{holistic token evaluation} in GlobalCom$^2$. While all three measures help preserve informative visual tokens to some extent, $s_i^{attn}$ shows notably different behavior from the other two metrics. Notably, $s_i^{sim}$ demonstrates performance closest to $s_i^{\texttt{[CLS]}}$ across various benchmarks, even outperforming it on MM-Vet. Based on these results, we adopt $s_i^{sim}$ as an alternative to $s_i^{\texttt{[CLS]}}$ for measuring information content of local crops and visual tokens when \texttt{[CLS]} token is unavailable, enabling training-free token compression for HR-LVLMs.

\section{Extension to Video Understanding}
\label{sec:videollms}

Given that VideoLLMs process sequential frames with substantial redundancy, analogous to HR-LVLMs with dynamic cropping, we extend GlobalCom$^2$ for efficient VideoLLMs. For video tokens $\mathbf{V} = \{\mathbf{V}_j\}_{j=1}^T$, we derive global representation $\mathbf{v}^g$ via global average pooling. We compute cosine similarity between token $i$ in frame $j$ and $\mathbf{v}^g$:
{\setlength\abovedisplayskip{2mm}
\setlength\belowdisplayskip{2mm}
\begin{equation}
    s_{j,i}^G = - \text{sim}(\mathbf{v}_{j,i}, \mathbf{v}^g) = -\frac{\mathbf{v}_{j,i} \cdot \mathbf{v}^g}{\|\mathbf{v}_{j,i}\| \|\mathbf{v}^g\|},
\end{equation}
}
where global score $s_{j,i}^G$ is negative cosine similarity, with lower similarity indicating higher distinctiveness. GlobalCom$^2$ calculates \textit{information richness} $s_j^G = \sum_{i \in \text{frame}_j} s_{j,i}^{G}$ per frame. Following Equations~\ref{eq:crop reletaive importance}-\ref{eq:crop ratio allocation}, this guides compression intensity ($r_j$), preserving more tokens in information-dense frames.
For local importance, GlobalCom$^2$ generates frame representation $\mathbf{v}_j^f$ through pooling and computes local scores $s_{j,i}^L$ as negative cosine similarity. Combining both perspectives per Equation~\eqref{eq:holistic token evalutation} yields \textit{holistic score} $s_{j,i}$. Video compression is:
{\setlength\abovedisplayskip{2mm}
\setlength\belowdisplayskip{2mm}
\begin{equation}
    \mathbf{V}_j \rightarrow \mathbf{\hat{V}}_j = \text{TopK}(\mathbf{V}_j, s_j,  r_j \times N).
\end{equation}
}
Thus, GlobalCom$^2$ enables efficient VideoLLM inference through frame-wise adaptive compression.

\section{Detailed Experimental Settings}
\label{sec:detailed setups}

\noindent \textbf{Benchmark Details.} We evaluate GlobalCom$^2$ on various multimodal understanding benchmarks detailed as follows:

\begin{itemize}
    \item \textbf{GQA}~\cite{Hudson:GQA}: Contains 113,018 real-world images from Visual Genome with structured scene graphs for compositional visual reasoning. Features 22M diverse questions with functional program representations, focusing on multi-hop reasoning and semantic understanding with balanced answer distributions to reduce dataset biases.
    
    \item \textbf{VizWiz}~\cite{Gurari:VizWiz}: Includes over 31,000 visual questions from blind users who took photos with phones and recorded spoken questions. Images are often poor quality due to lighting and framing issues, with conversational-style questions that may be unanswerable, representing real-world accessibility challenges.
    
    \item \textbf{SQA}~\cite{Lu:ScienceQA}: Contains 21,208 multimodal multiple-choice scientific questions from elementary and high school curricula, covering natural science, language science, and social science. Features 26 topics, 127 categories, and 379 different reasoning skills with detailed lectures (83.9\%) and explanations (90.5\%) for answers.
    
    \item \textbf{VQA\textsuperscript{T}}~\cite{Singh:TextVQA}: Features 28,408 high-resolution images from OpenImages with 45,336 questions focusing on reading and reasoning about text embedded in natural scenes. Requires OCR capabilities and text-based visual reasoning, with questions averaging 7.18 words in length.
    
    \item \textbf{AI2D}~\cite{kembhavi2016AI2D}: Contains 5,000 high-resolution scientific diagrams from textbooks paired with 16,000 multiple-choice questions requiring visual-spatial reasoning about complex scientific concepts, processes, and relationships depicted in educational illustrations.
    
    \item \textbf{MMStar}~\cite{chen2024:MMStar}: Contains 1,500 challenging samples across 6 core capabilities (coarse perception, fine-grained perception, instance interaction, logical reasoning, science \& technology, mathematics) designed to evaluate vision-indispensable reasoning without relying on shortcuts.
    
    \item \textbf{POPE}~\cite{Li:POPE}: Features 3,000 images with 9,000 binary yes/no questions specifically designed for detecting object hallucination phenomena in large vision-language models. Uses systematic adversarial evaluation to probe model reliability and factual accuracy.
    
    \item \textbf{MME}~\cite{Fu:MME}: Contains 2,374 high-resolution images across 14 perceptual and cognitive reasoning subtasks, including existence, count, position, color, OCR, commonsense reasoning, numerical calculation, text translation, and code reasoning. Uses yes/no questions for comprehensive evaluation.
    
    \item \textbf{MMB}~\cite{Liu:MMBench}: Features 2,974 multiple-choice questions designed for robust visual reasoning evaluation across 20 ability dimensions including object localization, attribute recognition, scene understanding, and spatial relationship reasoning.
    
    \item \textbf{MMB-CN}~\cite{Liu:MMBench}: Chinese counterpart of MMBench with 2,974 questions translated and culturally adapted for cross-lingual evaluation, maintaining the same 20 ability dimensions while incorporating Chinese cultural contexts and linguistic nuances.
    
    \item \textbf{MM-Vet}~\cite{Yu:MM-Vet}: Features 218 high-quality questions across 6 core VL capabilities (recognition, OCR, knowledge, language generation, spatial awareness, math) with complicated multi-modal reasoning chains requiring integration of multiple skills for comprehensive evaluation.
    
    \item \textbf{SEED-Bench}~\cite{Li2024:SEED-Bench}: Contains 19,242 human-annotated multiple-choice questions across 12 evaluation dimensions covering both image and video understanding tasks. Features hierarchical evaluation from basic perception to complex reasoning with balanced difficulty distribution.
    
    \item \textbf{MVBench}~\cite{li2024mvbench}: Defines 20 video understanding tasks requiring deep temporal comprehension beyond single-frame analysis, including dynamic scene understanding, temporal action localization, multi-object tracking, and causal reasoning across video sequences.
    
    \item \textbf{LongVideoBench}~\cite{wu2024longvideobench}: Focuses on long-context video understanding with 3,763 videos up to one hour duration and 6,678 questions across 17 categories including plot understanding, character analysis, temporal reasoning, and comprehensive video summarization.
    
    \item \textbf{MLVU}~\cite{zhou2024mlvu}: Features videos from 3 minutes to 2 hours with 9 tasks including topic reasoning, video summarization, needle-in-a-haystack retrieval, and ego-centric analysis, designed to evaluate long-form video comprehension and temporal memory capabilities.
    
    \item \textbf{VideoMME}~\cite{fu2024videomme}: Comprises 900 videos and 2,700 questions across six domains (Knowledge, Film \& Television, Sports, Life, Subtitles, Games) with durations from 11 seconds to 1 hour, featuring both with-subtitle and without-subtitle evaluation settings.
\end{itemize}

\begin{table*}[!t]    
\tablestyle{5pt}{1.0}

\setlength\tabcolsep{7.8pt}
\def\w{20pt} 
\scalebox{1.1}{
    \begin{tabular}{lccccccccc}
    \textbf{Method} & \textbf{VQAv2} & \textbf{GQA} & \textbf{SQA} & \textbf{VQA\textsuperscript{T}} & \textbf{POPE} & \textbf{MME} & \textbf{MMB} & \textbf{MMB-CN} & \textbf{Average} \\
    \shline
    \multicolumn{10}{l}{\textit{Upper Bound, 2880 Tokens}} \\
    LLaVA-NeXT-13B & 82.8 & 65.4 & 73.5 & 67.1 & 86.2 & 1575.9 & 70.0 & 64.2 & 100.0\% \\
    \hline
    \multicolumn{10}{l}{\textit{Ratio=75\%, Retain up to 2160 Tokens}} \\
    FasterVLM\tiny\texttt{(2024.12)} & \textbf{81.9} & 64.6  & 72.6 & 62.8  & 87.6  & 1560.3  & \textbf{69.8} & \textbf{64.8} & 98.6\% \\
     \textbf{GlobalCom$^2$ } & \textbf{81.9} & \textbf{65.0} & \textbf{72.8} & \textbf{65.2} & \textbf{87.8} & \textbf{1567.9} & 69.2  & 64.7  & \textbf{99.3\%} \\
    \hline
    \multicolumn{10}{l}{\textit{Ratio=50\%, Retain up to 1440 Tokens}} \\
    FasterVLM\tiny\texttt{(2024.12)} & \textbf{81.3} & 64.2  & 72.5 & 62.4  & 87.6  & 1534.1  & \textbf{69.5} & \textbf{64.1} & 97.5\% \\
     \textbf{GlobalCom$^2$ } & 81.0 & \textbf{64.7} & \textbf{73.2} & \textbf{64.6} & \textbf{87.7} & \textbf{1553.5} & 69.3  & 63.5  & \textbf{98.4\%} \\
    \hline
    \multicolumn{10}{l}{\textit{Ratio=25\%, Retain up to 720 Tokens}} \\
    FasterVLM\tiny\texttt{(2024.12)} & 78.9 & 62.3  & 72.1 & 61.2  & 86.1  & 1516.1  & 67.6  & 62.1  & 95.3\% \\
     \textbf{GlobalCom$^2$ } & \textbf{79.9} & \textbf{62.7} & \textbf{72.3} & \textbf{63.6} & \textbf{86.5} & \textbf{1531.2} & \textbf{67.9} & \textbf{62.2} & \textbf{96.1\%} \\
    \hline
    \multicolumn{10}{l}{\textit{Ratio=10\%, Retain up to 288 Tokens}} \\
    FasterVLM\tiny\texttt{(2024.12)} & 74.5 & 58.1  & 70.5 & 58.0  & 81.6  & 1386.2  & 61.7  & 53.5  & 88.6\% \\
     \textbf{GlobalCom$^2$ } & \textbf{77.0} & \textbf{58.3} & \textbf{71.8} & \textbf{60.3} & \textbf{82.4} & \textbf{1399.5} & \textbf{65.0} & \textbf{58.5} & \textbf{90.9\%} \\
    \end{tabular}%
    }
    \vspace{-1mm}
    \caption{\textbf{Comparison with FasterVLM on LLaVA-NeXT-13B across multiple benchmarks.}}
    \vspace{-2mm}
  \label{tab:main_results_13B}%
\end{table*}%

\noindent \textbf{Baseline Models.} We select LLaVA-NeXT~\cite{Liu:LLaVA-NeXT} and LLaVA-OneVision (SI)~\cite{li2024llava-ov} as our HR-LVLMs models, and follow the same inference
setting as the original paper as it is publicly available\footnote{LLaVA-NeXT: \url{https://github.com/haotian-liu/LLaVA/blob/main/docs/Evaluation.md}, LLaVA-OneVision: \url{https://github.com/LLaVA-VL/LLaVA-NeXT/blob/main/docs/LLaVA_OneVision.md}.}. LLaVA-NeXT and LLaVA-OneVision share a common three-component architecture: a pre-trained vision encoder, a large language model (LLM) backbone, and a two-layer MLP projector bridging the two. Specifically, LLaVA-NeXT employs CLIP-ViT-L-336px~\cite{Radford:CLIP} and Vicuna-v1.5 for vision and language modeling respectively, while LLaVA-OneVision utilizes SigLIP-So400m-Patch14-384~\cite{ZhaiM0B23:SigLIP} and Qwen2~\cite{yang2024Qwen2}. To effectively process high-resolution visual inputs, both models adopt flexible grid configurations - LLaVA-NeXT supports $\{2 \times 2, 1 \times \{2, 3, 4\}, \{2, 3, 4\} \times 1\}$ with maximum $5 \times 576$ grid tokens, while LLaVA-OneVision allows $\{1 \times 1, ... ,6 \times 6\}$ grids up to $10 \times 729$ tokens. For hyper-parameters, $\tau$ and $\alpha$ are respectively set as 10 and 0.5 for benchmark evaluations.

We also evaluate two VideoLLMs: LLaVA-OneVision and Qwen2-VL~\cite{Wang:Qwen2-VL}. LLaVA-OneVision unifies image and video tasks by encoding videos as long image-style token sequences, enabling strong zero-shot video understanding. Qwen2-VL adopts Naive Dynamic Resolution and Multimodal Rotary Position Embedding to achieve long video understanding ($>$20 min).

\noindent \textbf{Comparison Methods.} We compare our GlobalCom$^2$ with below dominant LVLM token compression methods: 

\begin{itemize}

    \item \textbf{FastV}~\cite{Chen:FastV} performs one-time pruning after the selected LLM layer based on attention weights between vision tokens and the last token.
    \item \textbf{PDrop}~\cite{Xing:PyramidDrop} introduces progressive token dropping using similar token selection metrics as FastV, forming a pyramid-like token structure that balances efficiency and performance.
    \item \textbf{SparseVLM}~\cite{Zhang:SparseVLM} ranks token importance by text-visual attention maps, pruning through pre-selected text prompts to reduce attention noise.
    \item \textbf{PruMerge}~\cite{Shang:LLaVA-PruMerge} integrates token pruning and merging by removing less important tokens using \texttt{[CLS]} attention weights with patch tokens and clustering retained tokens based on key similarity.
    
    \item \textbf{FasterVLM}~\cite{Zhang:FasterVLM} re-ranks visual tokens by using \texttt{[CLS]} attention scores with all patch tokens from ViT and preserves top-$k$ tokens.

\end{itemize}

We also compare with two VideoLLM-specific methods:

\begin{itemize}
    \item \textbf{DyCoke}~\cite{tao2024dycoke} groups video frames using a four-frame sliding window and performs temporal token merging within each window.
    \item \textbf{FrameFusion}~\cite{Fu2025:FrameFusion} adopts a two-stage strategy: first merging tokens across frames based on visual similarity, then selecting tokens within each frame based on importance.
\end{itemize}

All experiments in this work are conducted on NVIDIA A100-SXM4-80GB GPUs.

\section{Additional Experiments}
\label{sec:more exp}

\begin{table}[!t]
  \centering
  \setlength{\tabcolsep}{0.15pt}
    \scalebox{0.9}{\begin{tabular}{lcccccccc}
    \textbf{Method} & \textbf{GQA} & \textbf{VizWiz} & \textbf{SQA} & \textbf{AI2D} & \textbf{MMStar} & \textbf{MME} & \textbf{SEED} & \textbf{Avg.} \\
    \shline
    \multicolumn{6}{l}{\textit{Upper Bound, 7290 Tokens}} \\
    LLaVA-OV & 59.5 & 46.0 & 67.9 & 54.2 & 36.2 & 1216.1 & 63.8 & 100.0\% \\
    \hline
    $R=75\%$ & 58.9 & 44.5 & 67.9 & 52.7 & 37.1 & 1205.3 & 63.3 & 99.1\% \\
    $R=50\%$ & 57.4 & 44.3 & 67.6 & 52.0 & 37.0 & 1216.1 & 62.6 & 98.4\% \\
    \hline
    $R=25\%$ & 54.2 & 42.1 & 67.0 & 51.3 & 35.8 & 1201.8 & 60.4 & 95.4\% \\
    $R=15\%$ & 50.6 & 41.0 & 67.5 & 50.4 & 35.4 & 1140.2 & 56.8 & 91.1\% \\
    $R=10\%$ & 48.6 & 40.7 & 67.1 & 50.0 & 35.0 & 1085.7 & 54.9 & 90.5\% \\   
    \end{tabular}
    }
  \vspace{-2mm}
  \caption{\textbf{Results of GlobalCom$^2$ on LLaVA-OV-0.5B.}}
  \vspace{-2mm}
  \label{tab:llava-ov}%
\end{table}%


\begin{table*}[!t]
\centering
\tablestyle{5pt}{1.0}
\setlength\tabcolsep{6.4pt}
\scalebox{1.1}{
    \begin{tabular}{lcccccccc}
    \multirow{2}{*}{\textbf{Method}} & \multirow{2}{*}{\textbf{MVBench}} & \multirow{2}{*}{\textbf{LongVideoBench}} & \multirow{2}{*}{\textbf{MLVU}} & \multicolumn{4}{c}{\textbf{VideoMME}} & \multirow{2}{*}{\textbf{Average}} \\
        &  &  &  & \textbf{Overall} & \textbf{Short} & \textbf{Medium} & \textbf{Long} &  \\
    \shline
    \multicolumn{9}{l}{\textit{Upper Bound, 6272 Tokens}} \\
    LLaVA-OneVision-7B & 56.9 & 56.4 & 63.0 & 58.6 & 70.3 & 56.6 & 48.8 & 100.0\% \\
    \hline
    \multicolumn{9}{l}{\textit{Ratio=30\%, Retain 1882 Tokens}} \\
    DyCoke\tiny\texttt{(CVPR25)} & 56.6 & 54.7 & 60.3 & 56.1 & 67.1 & 54.6 & 46.6 & 96.5\% \\
    \hline
    \multicolumn{9}{l}{\textit{Ratio=25\%, Retain 1568 Tokens}} \\
    FastV\tiny\texttt{(ECCV24)} & 55.5 & 53.3 & 59.6 & 55.3 & 65.0 & 53.8 & 47.0 & 95.0\% \\
    PDrop\tiny\texttt{(CVPR25)} & 55.3 & 51.3 & 57.1 & 55.5 & 64.7 & 53.1 & 48.7 & 94.2\% \\
    SparseVLM\tiny\texttt{(ICML25)} & 56.4 & 53.9 & 60.7 & 57.3 & 68.4 & 55.2  & 48.1 & 97.5\% \\
    FrameFusion\tiny\texttt{(ICCV25)} & 56.0 & 54.8 & 61.7 & 57.5 & 68.2 & 55.7 & 48.6 & 98.1\% \\
    \textbf{GlobalCom$^2$} & \textbf{57.0} & \textbf{55.4} & \textbf{62.6} & \textbf{58.1} & \textbf{69.2} & \textbf{55.8} & \textbf{49.4} & \textbf{99.3\%} \\
    \hline
    \multicolumn{9}{l}{\textit{Ratio=15\%, Retain 941 Tokens}} \\
    FastV\tiny\texttt{(ECCV24)} & 51.6 & 48.3 & 55.0 & 48.1 & 51.4 & 49.4 & 43.3 & 85.0\% \\
    PDrop\tiny\texttt{(CVPR25)} & 53.2 & 47.6 & 54.7 & 50.1 & 58.7 & 48.7 & 45.0 & 87.4\% \\
    SparseVLM\tiny\texttt{(ICML25)} & 52.9 & 49.7 & 57.4 & 53.4 & 61.0 & 52.1 & 47.0 & 91.2\% \\
    \textbf{GlobalCom$^2$} & \textbf{54.2} & \textbf{52.9} & \textbf{60.7} & \textbf{55.8} & \textbf{66.0} & \textbf{53.3} & \textbf{48.1} & \textbf{95.3\%} \\
    \end{tabular}%
    }
    \vspace{-3mm}
    \caption{\textbf{Comparisons with LLaVA-OneVision across video understanding benchmarks.} Each benchmark spans different durations: MVBench (16s), LongVideoBench (1-60min), MLVU (3-120min), and VideoMME-S/M/L (1-3/3-30/30-60min).
    }
    \vspace{-5mm}
  \label{tab:main_results_video}%
\end{table*}%

\begin{table}[!t]
  \centering
  \setlength{\tabcolsep}{0.2pt}
  \scalebox{0.93}{
    \begin{tabular}{lcccccc}
    \multirow{2}{*}{\textbf{Method}} &  \multirow{2}{*}{\textbf{LongVideo.}} &  \multirow{2}{*}{\textbf{MLVU}}  &  \multicolumn{4}{c}{\textbf{VideoMME}} \\
    & & & \textbf{Overall} & \textbf{Short} & \textbf{Medium} & \textbf{Long} \\
    \shline
    \multicolumn{7}{l}{\textit{Upper Bound}} \\
    Qwen2-VL-7B & 56.0 & 60.4 & 57.6 & 70.0 & 54.6 & 48.2 \\
    \hline
    \multicolumn{7}{l}{\textit{Ratio=25\%}} \\
    FastV\tiny\texttt{(ECCV24)} & \multicolumn{6}{c}{Out of Memory (OOM)} \\
    DyCoke\tiny\texttt{(CVPR25)} & 50.5 & 55.2  & 51.5  &  62.1 & 47.1  & 45.1  \\
    \textbf{GlobalCom$^2$}  & \textbf{51.4} & \textbf{55.9} & \textbf{54.3} & \textbf{64.7}  & \textbf{50.1}  & \textbf{48.2} \\
    \end{tabular}%
    }
  \vspace{-3mm}
  \caption{\textbf{Comparisons on Qwen2-VL.} ``LongVideo'' is LongVideoBench. We conduct on 4 A100 GPUs.}
  \label{tab:qwen2_vl_video}%
  \vspace{-3mm}
\end{table}%

\subsection{Comparisons on LLaVA-NeXT-13B}
\label{subsec:13b comparisons}

Considering that both FasterVLM and GlobalCom$^2$ perform token compression at the vision encoding stage, Table~\ref{tab:main_results_13B} focuses on comparing these two methods on larger HR-LVLM LLaVA-NeXT-13B. The results reveal several key findings: \textbf{(i)} Our GlobalCom$^2$ outperforms FasterVLM on most benchmarks, maintaining \textbf{above 90\%} of uncompressed performance across different retention ratios on both LLaVA-NeXT-7B and 13B models. \textbf{(ii)} On visual text understanding (VQA\textsuperscript{T}), GlobalCom$^2$ shows particular strength at low retention ratios, surpassing FasterVLM by \textbf{2.3\%} at $R=10\%$, thanks to its effective preservation of object and textual information. \textbf{(iii)} While FasterVLM performs slightly better on general visual tasks (VQAv2, MMB, MMB-CN) at high retention ratios due to its uniform token compression, GlobalCom$^2$ demonstrates superior degradation resistance at low ratios. For example, when $R$ drops from 25\% to 10\%, FasterVLM's performance significantly decreases by 5.9\% and 8.6\% on MMB and MMB-CN, while GlobalCom$^2$ only drops by 2.9\% and 3.7\%, benefiting from our carefully designed ``global-to-local'' guided compression strategy.

\subsection{Results on LLaVA-OneVision-0.5B}
\label{subsec:llava-ov-0.5b}

We adopt $s_i^{sim}$ proposed in Section~\ref{sec:more non-cls discussions} as the measure for crop information richness and token informativeness, and evaluate our GlobalCom$^2$ on the LLaVA-OneVision model with more local crops across multiple benchmarks.

Results show that GlobalCom$^2$ maintains satisfactory performance even under low retention ratios ($R=25\%, 15\%, 10\%$). Particularly, GlobalCom$^2$ achieves promising results on AI2D and MMStar benchmarks which involve high-resolution images. Interestingly, we observe that on the SQA benchmark, the model performance shows minimal degradation as token retention rate decreases. We attribute this to SQA's lower dependency on visual signals and higher reliance on LLM capabilities. Given that LLaVA-OneVision employs the powerful Qwen2~\cite{yang2024Qwen2} as its LLM decoder, it consistently performs well on this benchmark regardless of retention ratio.

\subsection{Results on Video Understanding.} 

Table~\ref{tab:main_results_video} presents GlobalCom$^2$'s extension to video understanding with LLaVA-OneVision, demonstrating two key advantages: \textbf{(i) Superior VideoLLM performance:} GlobalCom$^2$ achieves \textbf{2.7\%} average improvement over VideoLLM-specific method DyCoke~\cite{tao2024dycoke} while using fewer vision tokens. \textbf{(ii) Advantages on long video understanding:} At $R=15\%$, GlobalCom$^2$ achieves \textbf{3.2} and \textbf{3.3} points higher than the second-best method on LongVideoBench and MLVU, demonstrating the effectiveness of our ``global-to-local'' guided compression design. Table~\ref{tab:qwen2_vl_video} shows GlobalCom$^2$ significantly outperforms DyCoke on Qwen2-VL, confirming its scalability. Additionally, FastV explicitly computes the full attention matrix in the certain layer, causing OOM issues with high token counts and limiting its practical applications.

\begin{table*}[!t]
  \centering
\tablestyle{2.5pt}{1.0}  
\setlength\tabcolsep{2.2pt}
\def\w{20pt} 
\scalebox{1}{
    \begin{tabular}{lccccc c c}
    \textbf{Method} & \textbf{TFLOPs↓} & \textbf{Peak Memory (GB)↓} & \textbf{KV-Cache (MB)↓} & \textbf{Prefill Time (ms)↓} & \textbf{Throughput (samples/s)↑} & \textbf{Performance↑} \\
    \shline
    \multicolumn{7}{l}{\textit{Upper Bound, 2880 Tokens}} \\
    LLaVA-NeXT-7B & 41.7 & 23.8 & 1536.0 & 170.7 & 2.5 & 1519.0 \\
    \hline
    \multicolumn{7}{l}{\textit{Ratio=75\%, Retain up to 2160 Tokens}} \\
     \textbf{GlobalCom$^2$} & 30.4 \downtiny{(↓27\%)} & 17.8 \downtiny{(↓25\%)} & 1126.4 \downtiny{(↓27\%)} & 119.9 \downtiny{(↓30\%)} & 3.2 \downtiny{(1.3$\times$)} & 1548.4 \\
    \hline
    \multicolumn{7}{l}{\textit{Ratio=50\%, Retain up to 1440 Tokens}} \\
     \textbf{GlobalCom$^2$} & 19.7 \downtiny{(↓53\%)} & 16.2 \downtiny{(↓32\%)} & 755.0 \downtiny{(↓51\%)} & 74.5 \downtiny{(↓57\%)} & 4.2 \downtiny{(1.7$\times$)} & 1552.9 \\
    \hline
    \multicolumn{7}{l}{\textit{Ratio=25\%, Retain up to 720 Tokens}} \\
     \textbf{GlobalCom$^2$} & 9.6 \downtiny{(↓77\%)} & 14.8 \downtiny{(↓39\%)} & 377.0 \downtiny{(↓76\%)} & 34.6 \downtiny{(↓80\%)} & 5.3 \downtiny{(2.1$\times$)} & 1493.5 \\
    \hline
    \multicolumn{7}{l}{\textit{Ratio=10\%, Retain up to 288 Tokens}} \\
     \textbf{GlobalCom$^2$} & 3.8 \downtiny{(↓91\%)} & 14.2 \downtiny{(↓40\%)} & 151.0 \downtiny{(↓90\%)} & 13.3 \downtiny{(↓92\%)} & 6.8 \downtiny{(2.7$\times$)} & 1365.5 \\
    \shline
    \multicolumn{7}{l}{\textit{Upper Bound, 2880 Tokens}} \\
    LLaVA-NeXT-13B & 80.0 & 36.7 & 2457.6 & 295.1 & 1.9 & 1575.9 \\
    \hline
    \multicolumn{7}{l}{\textit{Ratio=75\%, Retain up to 2160 Tokens}} \\
     \textbf{GlobalCom$^2$} & 58.7 \downtiny{(↓27\%)} &  34.1 \downtiny{(↓7\%)} & 1843.2 \downtiny{(↓25\%)} & 211.6 \downtiny{(↓28\%)} & 2.4 \downtiny{(1.3$\times$)} & 1567.9 \\
    \hline
    \multicolumn{7}{l}{\textit{Ratio=50\%, Retain up to 1440 Tokens}} \\
     \textbf{GlobalCom$^2$} & 38.3 \downtiny{(↓52\%)} &  30.1 \downtiny{(↓18\%)} & 1228.8 \downtiny{(↓50\%)} & 134.6 \downtiny{(↓54\%)} & 3.2 \downtiny{(1.7$\times$)} & 1553.5 \\
    \hline
    \multicolumn{7}{l}{\textit{Ratio=25\%, Retain up to 720 Tokens}} \\
     \textbf{GlobalCom$^2$} & 18.7 \downtiny{(↓77\%)} & 27.5 \downtiny{(↓25\%)} & 590.0 \downtiny{(↓76\%)} & 64.1 \downtiny{(↓78\%)} & 4.8 \downtiny{(2.4$\times$)} & 1531.2 \\
    \hline
    \multicolumn{7}{l}{\textit{Ratio=10\%, Retain up to 288 Tokens}} \\
     \textbf{GlobalCom$^2$} & 7.4 \downtiny{(↓91\%)} & 26.2 \downtiny{(↓29\%)} & 236.0 \downtiny{(↓90\%)} & 25.0 \downtiny{(↓92\%)} & 5.8 \downtiny{(3.1$\times$)} & 1399.5 \\
    \end{tabular}%
   }
   \vspace{-2mm}
  \caption{\textbf{Comprehensive efficiency analysis with LLaVA-NeXT-7B/13B on one NVIDIA A100-SXM4-80GB GPU.}}
  \label{tab:detailed_efficiency}%
  \vspace{-2mm}
\end{table*}%


\subsection{Sensitivity Analysis of Hyper-parameters} 
\label{subsec:hyper-parameters}

\begin{figure*}[!t]
  \centering
   \includegraphics[width=\linewidth]{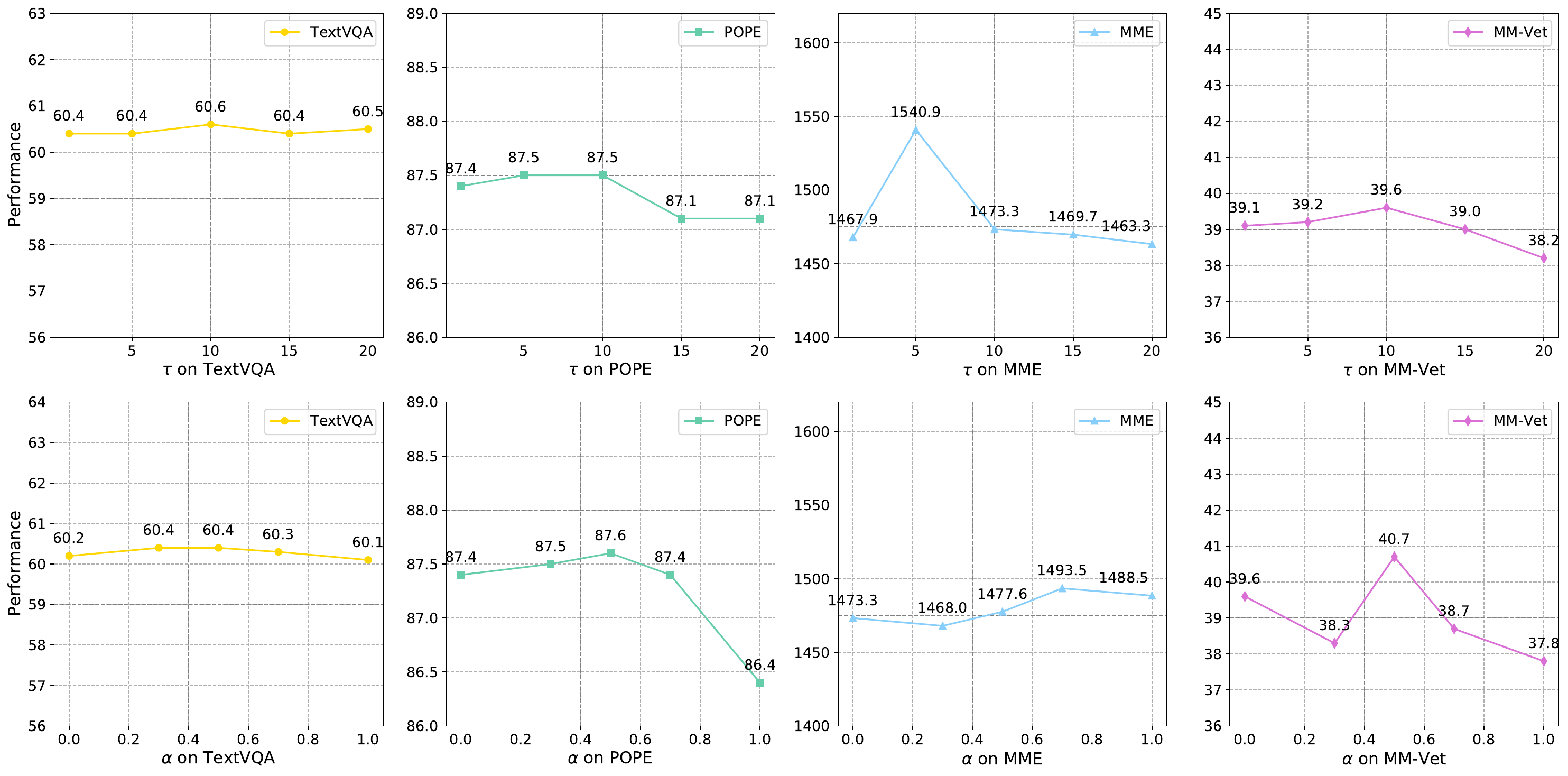}
   \vspace{-7mm}
   \caption{\textbf{Hyper-parameter sensitivity analysis of $\tau$ and $\alpha$.} Hyper-parameter $\tau$ controls the sharpness of probability distribution in softmax when assessing local crop informativeness. Hyper-parameter $\alpha$ determines the token retention threshold where higher values indicate stronger reliance on global guidance.} 
   
   \label{fig:hyper_para}
   \vspace{-2mm}
\end{figure*}

We further explore the hyper-parameter configurations $\tau$ and $\alpha$ of our GlobalCom$^2$ in Figure~\ref{fig:hyper_para}. 

Hyper-parameter $\tau$ serves as the temperature in the softmax function when assessing local crop significance, controlling the ``sharpness'' of probability distribution. A smaller $\tau$ leads to a sharper distribution, amplifying the differences in global visual importance among local crops. As shown in the first row of Figure~\ref{fig:hyper_para}, our GlobalCom$^2$ demonstrates robust performance across most benchmarks, particularly on VQA\textsuperscript{T}~\cite{Singh:TextVQA} and POPE~\cite{Li:POPE}, with varying $\tau$ validating our design principle for HR-LVLMs of allocating retention ratios based on each local crop's global importance. We empirically set $\tau=10$ to achieve optimal performance across most benchmarks.

Hyper-parameter $\alpha$ determines the criterion for token retention in local crops, where a larger $\alpha$ indicates stronger dependence on global guidance for token retention in local crops. Similar to $\tau$, the second row of Figure~\ref{fig:hyper_para} shows that our GlobalCom$^2$ exhibits consistent effectiveness across most benchmarks under different $\alpha$ settings, substantiating our dual-perspective token retention strategy of local crops by considering their significance at both global and local levels for HR-LVLMs with dynamic cropping.

\subsection{Detailed Efficiency Analysis}
\label{subsec:detailed efficiency}

We conduct a comprehensive analysis of both theoretical and practical efficiency of our GlobalCom$^2$ with LLaVA-NeXT-7B/13B on a single NVIDIA A100-SXM4-80GB GPU, as shown in Table~\ref{tab:detailed_efficiency}. We measure computational efficiency using TFLOPs and peak memory consumption directly, while other metrics are estimated using LLM-Viewer~\cite{Yuan:LLM-Viewer}. Given that the sequence length of visual tokens substantially exceeds that of textual and system tokens, we \textit{exclude} the latter two from our theoretical analysis. For practical efficiency, ``Throughout'' and the corresponding ``Performance'' measurements are conducted on the MME~\cite{Fu:MME}, which includes 2374 examples.

In Table~\ref{tab:detailed_efficiency}, GlobalCom$^2$ substantially improves the computational efficiency of LLaVA-NeXT models across both theoretical and practical metrics. Specifically, while preserving model performance, our method achieves significant reductions in GPU memory consumption and notable acceleration in inference speed compared to the original model. Most importantly, these efficiency improvements are achieved through a training-free compression scheme, demonstrating its practical value.

\subsection{More Visualizations of Token Compression}
\label{subsec:More Visualizations}

\begin{figure*}[!t]
  \centering
   \includegraphics[width=\linewidth]{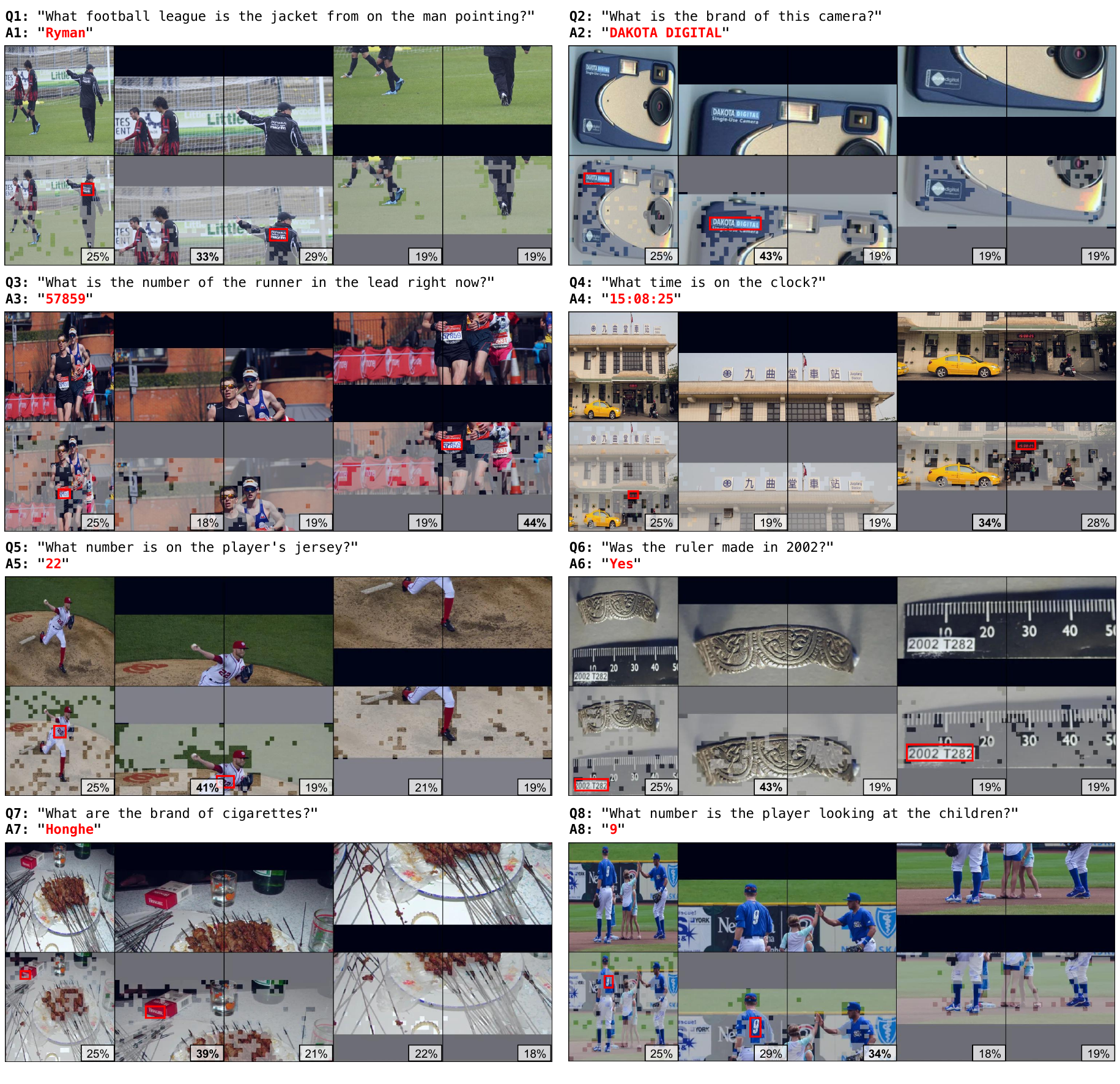}
   \vspace{-7mm}
    \caption{\textbf{More visualization of token compression by GlobalCom$^2$.} The presented examples are from VQA\textsuperscript{T}, where grey masks indicate discarded visual tokens.}

   \label{fig:more_compression_performance}
   \vspace{-2mm}
\end{figure*}

In Figure~\ref{fig:more_compression_performance}, we present more token compression visualizations of GlobalCom$^2$. It is clear that in all cases, \textit{entity-rich} regions are effectively preserved while redundant ones are removed. For crops with high visual redundancy (\textit{e.g.}, the camera's golden casing in the top-right case), it applies aggressive compression ($r_4 = 19\%$). Conversely, for semantically dense regions (\textit{e.g.}, the text-rich first crop in the top-right case), it maintains more tokens ($r_1 = 43\%$) to ensure detailed understanding. More cases in Appendix~\ref{fig:more_compression_performance} further demonstrate GlobalCom$^2$'s consistent ability to adaptively preserve significant regions while removing redundancy at both global and local views.

\section{Algorithm Illustration}\label{sec:algorithm}

We present a comprehensive description of GlobalCom$^2$ token compression for both global thumbnail and local crops in Algorithm~\ref{alg:global_compression} and Algorithm~\ref{alg:local_compression}, respectively.

\clearpage
\begin{algorithm}[t]
\caption{GlobalCom$^2$: Thumbnail Compression}
\label{alg:global_compression}
\begin{algorithmic}[1]
\Require  
Global thumbnail tokens $\mathbf{X}^G \in \mathbb{R}^{N \times D}$, 
Vision encoder, 
Preset retention ratio $R \in (0,1]$
\Ensure Compressed tokens $\mathbf{\hat{X}}^G \in \mathbb{R}^{k \times D}$ ($k=R\cdot N$)

\State // For models with \texttt{[CLS]} token
\If{model has \texttt{[CLS]} token}
    \State Get \texttt{[CLS]} query $\mathbf{q}^{\texttt{CLS}}$ and Key matrix $\mathbf{K}$
    \State Compute importance scores:
    \[
    s_i^G \leftarrow \frac{\exp(\mathbf{q}^{\texttt{CLS}} \mathbf{K}_i^\top / \sqrt{D})}{\sum_{j=1}^N \exp(\mathbf{q}^{\texttt{CLS}} \mathbf{K}_j^\top / \sqrt{D})}, \forall i
    \]
\Else
    \State // For models without \texttt{[CLS]} token
    \State Compute global mean vector: $\mathbf{g} \leftarrow \frac{1}{N}\sum_{i=1}^N \mathbf{x}_i$
    \State Compute cosine similarity scores:
    \[
    s_i^G \leftarrow -\frac{\mathbf{x}_i \cdot \mathbf{g}}{\|\mathbf{x}_i\| \|\mathbf{g}\|}, \forall i
    \]
\EndIf

\State Sort indices: $\text{idx} \leftarrow \text{argsort}([s_1^G,...,s_N^G])$ (descending)
\State Retain top-$k$ tokens: $\mathbf{\hat{X}}^G \leftarrow \mathbf{X}^G[\text{idx}[1:k]]$

\State \Return $\mathbf{\hat{X}}^G$
\end{algorithmic}
\end{algorithm}

\begin{algorithm}[t]
\caption{GlobalCom$^2$: Local Crop Compression}
\label{alg:local_compression}
\begin{algorithmic}[1]
\Require  
Local crops $\{\mathbf{X}_j^L\}_{j=1}^n$,
Global importance scores $s^G$,
Preset retention ratio $R$,
Balance weight $\alpha$,
Temperature $\tau$
\Ensure Compressed crop tokens $\{\mathbf{\hat{X}}_j^L\}_{j=1}^n$

\State \textbf{Adaptive Compression Adjustment:}
\For{each crop $j \in [1,n]$}
    \State Compute information richness:
    \[
    s_j^G \leftarrow \sum_{i \in \text{crop}_j} s_i^G
    \]
    \State Normalize: $\tilde{s}_j \leftarrow (s_j^G - \max\{s_j^G\})/\tau$
\EndFor

\State Compute importance weights: 
\[
\sigma_j \leftarrow \frac{\exp(\tilde{s}_j)}{\sum_{l=1}^n \exp(\tilde{s}_l) + \epsilon}
\]

\State Adjust retention ratios: 
\[
r_j \leftarrow R \times (1 + \sigma_j - \frac{1}{n})
\]

\State \textbf{Holistic Token Evaluation:} 
\For{each crop $j \in [1,n]$}
    \State Compute local scores $s_j^L$ via vision encoder
    \State Get global sub-scores $\hat{s}_j^G$ via interpolation
    \State L2-normalize both $s_j^L$ and $\hat{s}_j^G$
    \State Combine scores: $s_{j,i} \leftarrow \alpha \hat{s}_{j,i}^G + (1-\alpha)s_{j,i}^L$
    \State Compress: $\mathbf{\hat{X}}_j^L \leftarrow \text{TopK}(\mathbf{X}_j^L, s_j, r_j \times N)$
\EndFor

\State \Return $\{\mathbf{\hat{X}}_j^L\}_{j=1}^n$
\end{algorithmic}
\end{algorithm}

\end{document}